%% file: paper.tex
\documentclass[opre,nonblindrev]{informs3} 

\OneAndAHalfSpacedXI 

\usepackage{natbib}
 \bibpunct[, ]{(}{)}{,}{a}{}{,}%
 \def\bibfont{\small}%
\newcommand{\blue}[1]{#1}

\usepackage{graphicx, url, color}

\usepackage{enumitem}
\usepackage{times}
\usepackage{algorithm}
\usepackage{algorithmic}
\usepackage[citecolor=blue, linkcolor=blue, colorlinks=true]{hyperref}
\usepackage{etoolbox}

\newtheorem{assumption}{Assumption}

\newtheorem{lem}{Lemma}
\newtheorem{thm}{Theorem}

\newtheorem{prop}{Proposition}

\newtheorem{rem*}{Remark}

\global\long\def\indic{\mathbb{I}}
\global\long\def\mA{\mathcal{A}}
\global\long\def\mS{\mathcal{S}}

\global\long\def\mP{\mathcal{P}}
\global\long\def\mR{\mathcal{R}}
\global\long\def\mF{\mathcal{F}}

\global\long\def\P{\Pr}
\global\long\def\E{\mathbb{E}}
\global\long\def\P{\mathbb{P}}

\global\long\def\emcts{\varepsilon_{\text{mcts}}}
\global\long\def\tsl{\theta_{\text{sl}}}
\global\long\def\esl{\varepsilon_{\text{sl}}}

\ECRepeatTheorems

\EquationsNumberedThrough    

\begin{document}


\RUNAUTHOR{Shah, Xie, and Xu}

\RUNTITLE{MCTS, Non-stationary MAB}

\TITLE{Non-Asymptotic Analysis of Monte Carlo Tree Search}

\ARTICLEAUTHORS{%
\AUTHOR{Devavrat Shah}
\AFF{LIDS, Massachusetts Institute of Technology, Cambridge, MA 02139, \EMAIL{devavrat@mit.edu}} 
\AUTHOR{Qiaomin Xie}
\AFF{ORIE, Cornell University,  Ithaca, NY 14853, \EMAIL{qiaomin.xie@cornell.edu}}
\AUTHOR{Zhi Xu}
\AFF{LIDS, Massachusetts Institute of Technology, Cambridge, MA 02139, \EMAIL{zhixu@mit.edu}}
} 

\ABSTRACT{%
In this work, we consider the popular tree-based search strategy within the framework of reinforcement learning,
the Monte Carlo Tree Search (MCTS)\blue{,} in the context of infinite-horizon discounted cost Markov Decision Process (MDP). 
While MCTS is believed to provide an approximate value function for a given state with enough simulations, cf. \cite{kocsis2006bandit,kocsis2006improved}, the claimed proof  of this 
property is incomplete. This is due to the fact that the variant of MCTS, the Upper Confidence \blue{B}ound for Trees (UCT), 
analyzed in prior works utilizes ``logarithmic'' bonus term for balancing exploration and exploitation within the tree-based 
search, following the insights from stochastic multi-arm bandit (MAB) literature, cf. \cite{ucb1, ucb2}. In effect, such 
an approach assumes that the regret of \blue{the} underlying recursively dependent non-stationary MABs concentrates around 
their mean exponentially in the number of steps, which is unlikely to hold as pointed out in \cite{audibert2009exploration}\blue{,} 
even for stationary MABs. 

As the key contribution of this work, we establish polynomial concentration property of regret for a class of \emph{non-stationary} 
Multi-Arm Bandits (MAB). This in turn establishes that the MCTS with appropriate \emph{polynomial} rather than \emph{logarithmic} 
bonus term in UCB has the claimed property of  \cite{kocsis2006bandit,kocsis2006improved}. Interestingly enough, empirically 
successful approaches \blue{(}cf. \cite{silver2017mastering}\blue{)} utilize a similar polynomial form of MCTS as suggested by our result. 
Using this as a building block, we argue that MCTS, combined with nearest neighbor supervised learning\blue{,} acts as a ``policy 
improvement" operator, i.e., it iteratively improves value function approximation for \emph{all} states, due to combining with 
supervised learning, despite evaluating at only finitely many states. In effect, we establish that to learn an $\varepsilon$ 
approximation of the value function with respect to $\ell_\infty$ norm, MCTS combined with nearest neighbor requires a sample 
size scaling as $\widetilde{O}\big(\varepsilon^{-(d+4)}\big)$, where $d$ is the dimension of the state space. This is nearly 
optimal due to a minimax lower bound of $\widetilde{\Omega}\big(\varepsilon^{-(d+2)}\big)$\blue{,} suggesting \blue{the} strength of the variant 
of MCTS we propose here and our resulting analysis.\footnote{{An extended abstract of this paper is accepted for presentation at ACM SIGMETRICS 2020.}}
}%



\maketitle

%


\input{introduction}
\input{preliminary}

\input{mctsalgo}

\input{rlalgo}
\input{nonstatmab}

\input{proof_nonstatmab}

\input{proof_mcts}

\input{proof_sl}

\section{Conclusion} \label{sec:discussion}

In this paper, we introduce a {\em correction} of the popular Monte Carlo Tree Search (MCTS) policy for improved value function estimation for a given state, using an existing value function estimation for the entire state space. This correction was obtained through careful, rigorous analysis of a non-stationary Multi-Arm Bandit where rewards are dependent and non-stationary. In particular, we analyzed a variant of the classical Upper Confidence Bound policy for such an MAB. Using this as a building block, we establish rigorous performance guarantees for the {\em corrected} version of MCTS proposed in this work. This, to the best of our knowledge, is the first mathematically correct analysis of the UCT policy despite its popularity since it has been proposed in 
literature \citep{kocsis2006bandit,kocsis2006improved}. We further establish that the proposed MCTS policy, when combined with nearest neighbor supervised learning, leads to near optimal sample complexity for obtaining estimation of value function within a given tolerance, where the optimality is in the minimax sense. This suggests the tightness of our analysis as well as the utility of the MCTS policy. 

We take a note that much of this work was inspired by the success of AlphaGo Zero (AGZ) which utilizes MCTS combined with supervised learning. Interestingly enough, the correction of MCTS suggested by our analysis is qualitatively similar to the version of MCTS utilized by AGZ as reported in practice. This seeming coincidence may suggest further avenue for practical utility of versions of the MCTS proposed in this work and is an interesting direction for future work.

\bibliographystyle{informs2014}
\bibliography{mcts}

\begin{APPENDICES}

\input{Appendix_lower_bound.tex}

\end{APPENDICES}

\end{document}

%% file: introduction.tex
\section{Introduction} \label{sec:intro}


Monte Carlo Tree Search (MCTS) is a search framework for finding optimal decisions, based on the search tree built by random sampling of the decision space~\citep{browne2012survey}. MCTS has been widely used in sequential decision makings that have a tree representation, exemplified by games and planning problems.  Since MCTS was first introduced, many variations and enhancements have been proposed. Recently, MCTS has been combined with deep
neural networks for reinforcement learning, achieving remarkable success for games of Go~\citep{silver2016go,silver2017mastering}, chess and shogi~\citep{silver2017chess}. In particular, AlphaGo Zero (AGZ)~\citep{silver2017mastering} 
employs supervised learning to learn a policy/value function (represented by a neural network) based on samples generated via MCTS; the neural network is recursively used to estimate the value of leaf nodes in the next iteration of MCTS for simulation guidance.




Despite the wide application and empirical success of MCTS, there is only limited work on theoretical guarantees of MCTS and its variants. 
One exception is the work of~\cite{kocsis2006bandit} and \cite{kocsis2006improved}, which propose running tree search by applying the Upper 
Confidence Bound algorithm --- originally designed for stochastic multi-arm bandit (MAB) problems \citep{ucb1, ucb2} --- to each node of the tree. This leads to the 
so-called UCT (Upper Confidence Bounds for Trees) algorithm, which is one of the  popular forms of MCTS. In 
\cite{kocsis2006bandit}, certain asymptotic optimality property of UCT is claimed. The proof therein is, however, incomplete, 
as we discuss in greater details in Section~\ref{subsec:related_work}. More importantly, UCT as suggested in 
\cite{kocsis2006bandit} requires exponential concentration of regret for the underlying non-stationary MAB, 
which is unlikely to hold in general even for stationary MAB as pointed out in \cite{audibert2009exploration}.  

Indeed, rigorous analysis of MCTS is subtle, even though its asymptotic convergence may seem natural. A key challenge is that 
the tree policy (e.g., UCT) for selecting actions typically needs to balance exploration and exploitation, so the random sampling process 
at each node is non-stationary (non-uniform) across multiple simulations. A more severe difficulty arises due to the hierarchical/iterative 
structure of tree search, which induces complicated probabilistic dependency between a node and the nodes within its sub-tree. {Specifically, as part of simulation within MCTS, at each intermediate node (or state), the action is chosen based on the outcomes of the past simulation steps within the sub-tree of the node in consideration. 
}
{Such strong dependencies {across time (i.e., depending on the history) and space (i.e., depending on the sub-trees downstream)} among nodes makes the analysis non-trivial. }

The goal of this paper is to provide a rigorous theoretical foundation for MCTS.  In particular, we are interested in the following: 
\begin{itemize}

\item What is the appropriate form of MCTS for which the asymptotic convergence property claimed in the literature (cf.\ \cite{kocsis2006bandit, kocsis2006improved}) holds?

\item Can we rigorously establish the ``strong policy improvement'' property of MCTS when combined with supervised learning as observed in the literature (e.g., in \cite{silver2017mastering})? If yes, what is the quantitative form of it?
 
\item Does \blue{supervised} learning combined with MCTS lead to the optimal policy, asymptotically? If so, what is its finite-sample (non-asymptotic) performance?

\end{itemize}

\subsection{Our contributions} 
\label{sec:contributions}

As the main contribution of this work, we provide affirmative answers to all of the above questions. In what follows, we 
provide \blue{a} brief overview of our contributions and results. 

\medskip
\noindent {\bf Non-stationary MAB and recursive polynomial concentration.} In 
stochastic Multi Arm Bandit (MAB), the goal is to discover amongst finitely many actions (or arms), 
the one with the best average reward while choosing as few non-optimal 
actions as possible in the process. The rewards for any given arm are assumed to be 
independent and identically distributed (i.i.d.). The usual exponential concentration for 
such i.i.d. and hence stationary processes leads to UCB algorithm with \blue{a} {\em logarithmic} bonus
term: at each time, choose action with maximal index (ties broken arbitrarily), where 
the index of an arm is defined as the empirical mean reward plus constant times $\sqrt{\log t/ s}$, where $t$ is the total number of trials so far and $s \leq t$ is the number of times 
the particular action is chosen in these $t$ trials. 

The goal in the Monte Carlo Tree Search (MCTS) is very similar to the MAB setup described above -- choose an action at a given query state that gives the best average reward. However, the reward
depends on the future actions. Therefore, to determine the best action for the given state, one has 
to take future actions into account and MCTS does this by simulating future via
effectively expanding
all possible future actions recursively in the form of (decision-like) \blue{trees}. In essence, the optimal
action at the root of such a tree is determined by finding optimal path in the tree. And determining
this optimal path requires solving multiple MABs, one per each intermediate node within the tree. 
Apart from the MABs associated with the lowest layer of the tree, all the MABs associated with the intermediate 
nodes turn out to have rewards that are the rewards generated by MAB algorithms for nodes downstream. 
This creates complicated, hierarchically inter-dependent MABs. 

To determine the appropriate, UCB-like index algorithm for each node of the MCTS tree, it is essential to understand the concentration property of the rewards, i.e., concentration of regret for MABs associated with nodes downstream. While the rewards at leaf level may enjoy exponential concentration due to independence, the regret of any algorithm even for such an MAB is unlikely to have exponential concentration in general, cf.\ \cite{audibert2009exploration, salomon2011deviations}. Further, the MAB of our interest has non-stationary rewards due to strong dependence across hierarchy. Indeed, an oversight of this complication led \cite{kocsis2006bandit,kocsis2006improved} to suggest UCT 
inspired by the standard UCB algorithm for MABs with stationary, independent rewards. 

As an important contribution of this work, we formulate an appropriate form of non-stationary
MAB which correctly models the MAB at each of the node in the MCTS tree. For such a non-stationary
MAB, we define UCB algorithm with appropriate index and under which we establish appropriate concentration of the induced regret. This, in turn, allows us to 
recursively define the UCT algorithm for MCTS by appropriately defining index for each of the node-action within the MCTS tree. 
Here we provide a brief summary. 

Given $[K]=\{1,\dots, K\}$ actions or arms, let $X_{i,t}$ denote the reward generated by playing arm $i \in [K]$ for the $t$th time. Let empirical mean over $n$ trials for arm $i$ be $\bar{X}_{i,n}=\frac{1}{n}\sum_{t=1}^{n}X_{i,t}$, and let $\mu_{i,n}=\E[\bar{X}_{i,n}]$ be its expectation. Suppose $\mu_{i,n}\to \mu_i$ as $n\to \infty$ for all $i \in [K]$ and let there exist constants, $\beta>1$, $\xi>0$, and $1/2\leq\eta<1$ such that for every $z\geq 1$ and every integer $n\geq 1$,
\begin{align*}
\P\big(| n\bar{X}_{i,n}-n\mu_{i} | \geq n^\eta z\big) & \leq\frac{\beta}{z^{\xi}}.
\end{align*}
Note that for i.i.d.\ bounded rewards, above holds for $\eta = \blue{1/2}$ for any finite $\xi$ due to exponential concentration. 
We propose to utilize the UCB algorithm where at time $t$, the arm $I_t$ is chosen  according to
\begin{equation}
I_{t}\in\arg\max_{i\in[K]}\big\{\bar{X}_{i,T_{i}(t-1)}+B_{t-1,T_{i}(t-1)}\big\},\label{eq:app_non_ucb.re}
\end{equation}
where $T_i(t)=\sum_{l=1}^t\mathbb{I}{\{I_l = i\}}$ is the number of times arm $i$ has been played, up to (including) time $t$, and the bias or bonus term $B_{t,s}$ is defined as
\begin{equation*}
B_{t,s}=\frac{  \beta^{1/\xi}\cdot t^{\eta(1-\eta)}}{s^{1-\eta}}.
\end{equation*} 
Let $\mu_*=\max_{i\in[K]}\mu_i$ \blue{and let} $\bar{X}_{n}$
denote the empirical average of the rewards collected.
Then, we establish that 
$\E[\bar{X}_{n}]$ \blue{converges to} $\mu_{*}$, and \blue{that} for every $n\geq 1$ and every $z\geq 1$, a similar polynomial concentration holds:
\begin{align*}
\P\big(|n\bar{X}_{n}-n\mu_{*}| \geq n^{\eta} z\big) & \leq\frac{\beta'}{z^{\xi'}},
\end{align*}
where $\xi' = \xi \eta(1-\eta) - 1$, and $\beta'>1$ \blue{is} a large enough constant.  The precise statement can be found as 
Theorem \ref{thm:non.stat.mab} in Section \ref{sec:nonstatmab}.

\medskip
\noindent {\bf Corrected UCT for MCTS and non-asymptotic analysis.}
For MCTS, as discussed above, the leaf nodes have rewards that can be viewed as generated per
standard stationary MAB. Therefore, the rewards for each arm (or action) at the leaf level in MCTS 
satisfy the required concentration property with $\eta = 1/2$
due to independence.   \blue{Hence}, from our result for non-stationary MAB, we 
immediately obtain that we can recursively apply the UCB algorithm per \eqref{eq:app_non_ucb.re} 
at each level in the MCTS with $\eta = 1/2$ and appropriately adjusted constants $\beta$ and $\xi$. 
In effect, we obtain modified UCT where the bias or bonus term  $B_{t,s}$ scales as $t^{1/4}/s^{1/2}$. 
This is in constrast to $B_{t,s}$ scaling as $\sqrt{\log t / s}$ in the standard UCB as well as UCT suggested 
in the literature, cf. \cite{kocsis2006bandit, kocsis2006improved}. 

By recursively applying the convergence and concentration property of the non-stationary MAB 
for the resulting algorithm for MCTS, we establish that for any query state $s$ of the MDP, using
$n$ \blue{simulations} of the MCTS, we can obtain \blue{a} value function \blue{estimation} within \blue{error}
$\delta \varepsilon_{0}+O\big(n^{-1/2}\big)$, \blue{if we start with a value function \blue{estimation}  for all the leaf 
nodes within error $\varepsilon_0$ for some $\delta < 1$} (independent of $n$, dependent on depth of MCTS tree). 
That is, MCTS is indeed asymptotically correct as was conjectured in the prior literature. For details, 
see Theorem \ref{thm:MCTS} in Section \ref{sec:mcts}.

\medskip
\noindent {\bf MCTS with supervised learning, strong policy improvement, and near optimality.}
The result stated above \blue{for} MCTS implies its ``bootstrapping'' property -- if we start with a value function 
estimation for {\em all} state within error $\varepsilon$, then MCTS can produce estimation of value function
for a {\em given query} state within error \blue{less than} $ \varepsilon$ with enough simulations. By coupling such 
improved estimations of value function for a number of query states, combined with expressive enough
supervised learning, one can hope to generalize such improved estimations of value function for {\em all}
states. That is, MCTS coupled with supervised learning can be ``strong policy improvement operator''. 

Indeed, this is precisely what we establish by utilizing nearest neighbor supervised learning. Specifically, we establish that with $\widetilde{O}\big(\frac{1}{\varepsilon^{4+d}}\big)$ number of samples, MCTS with nearest neighbor finds an $\varepsilon$ approximation of the optimal value function with respect to $\ell_\infty$-norm; here $d$ is the dimension of the state space. This is nearly optimal in view of a minimax lower bound of $\widetilde{\Omega}\big(\frac{1}{\varepsilon^{2+d}}\big)$  \citep{shah2018qlnn}. For details, 
see Theorem \ref{thm:MCTS_SL_deterministic} in Section \ref{sec:RL}.

\medskip
\noindent {\bf An Implication.}
As mentioned earlier, the modified UCT policy per our result suggests using bias or bonus
term $B_{t,s}$ that scales as $t^{1/4}/s^{1/2}$ at each node within the MCTS. Interestingly enough, 
the empirical results of AGZ are obtained by utilizing $B_{t,s}$ that scales as $t^{1/2}/s$. This is qualitatively similar to what our results suggests and in contrast to the classical UCT.

\subsection{Related work} \label{subsec:related_work}

Reinforcement learning aims to approximate the optimal value function and policy directly from experimental data. A variety of algorithms have been developed, including model-based approaches, model-free approaches like tabular Q-learning ~\citep{watkins1992q}, and parametric approximation such as linear architectures~\citep{sutton1988learning}. More recent work approximates the value function/policy by deep neural networks~\citep{mnih2015human,schulman2015trust,schulman2017proximal,silver2017mastering,yang2019harnessing}, which can be trained using temporal-difference learning or Q-learning~\citep{van2016deep,mnih2016asynchronous,mnih2013playing}. 

MCTS is an alternative approach, which as discussed\blue{,} estimate\blue{s} the (optimal) value of states by building a search tree from Monte-Carlo simulations~\citep{kocsis2006bandit,chang2005adaptive,coulom2006efficient,browne2012survey}. \cite{kocsis2006bandit} and \cite{kocsis2006improved} argue for the asymptotic convergence of MCTS with standard UCT. However, the proof is incomplete~\citep{szepesv_2019_personal}. 
A key step towards proving the claimed result is to show the convergence and concentration properties of the regret for UCB 
under non-stationary reward distributions. In particular, to establish an exponential concentration of regret (Theorem 5, \citep{kocsis2006improved}), Lemma 14 is applied. However, it requires conditional independence of $\{Z_i\}$ sequence, which does not hold, hence making the conclusion of 
exponential concentration questionable. Therefore, the proof of the main result (Theorem~7, \citep{kocsis2006improved}), which applies Theorem 5 with an inductive argument, is incorrect as stated. 

In fact, it may be infeasible to prove Theorem 5 in \citep{kocsis2006improved} as it was stated. 
For example, the work of \cite{audibert2009exploration} shows 
that for bandit problems, the regret under UCB concentrates around its expectation polynomially and {\em not exponentially} as desired in \cite{kocsis2006improved}. 
Further, \cite{salomon2011deviations} prove that for any strategy that does not use the knowledge of time horizon, it is infeasible to improve this polynomial concentration
and establish exponential concentration. Our result is consistent with these fundamental bound of stationary MAB --- we establish polynomial concentration 
of regret for non-stationary MAB, which plays a crucial role in our analysis of MCTS. Also see the work \cite{munos2014bandits} for a discussion of 
the issues with logarithmic bonus terms for tree search.

While we focus on UCT in this paper, we note that there are other variants of MCTS developed for a diverse range of applications. The work of \cite{coquelin2007bandit} introduces flat UCB in order to improve the worst case regret bounds of UCT. \cite{schadd2008singleMCTS} modifies MCTS for single-player games by adding to the standard UCB formula a term that captures the possible deviation of the node. In the work by \cite{sturtevant2008UCT}, a variant of MCTS is introduced for multi-player games by adopting the max$^n$ idea. In addition to turn-based games like Go and Chess, MCTS has also been applied to real-time games (e.g., Ms.\ PacMan, Tron and Starcraft) and nondeterministic games with imperfect information. The applications of MCTS go beyond games, and appear in areas such as optimization, scheduling and other decision-making problems. We refer to the survey on MCTS by~\cite{browne2012survey} for other variations and applications.

It has become popular recently to combine MCTS  with deep neural networks, which serve to approximate the value function and/or policy~\citep{silver2016go,silver2017mastering,silver2017chess}. For instance, in AlphaGo Zero, MCTS uses the neural network to query the value 
of leaf nodes for simulation guidance; the neural network is then updated with sample data generated by MCTS-based policy and used in tree search in the next iteration. \cite{azizzadenesheli2018gats} develop generative adversarial tree search that generates roll-outs with a learned GAN-based dynamic model 
and reward predictor, while using MCTS for planning over the simulated samples and a deep Q-network to query the Q-value of leaf nodes. 

In terms of theoretical results, the closest work to our paper is~\cite{jiang2018feedback}, where they also consider a batch, MCTS-based 
reinforcement learning algorithm, which is a variant of AlphaGo Zero algorithm. The key algorithmic difference from ours lies in the leaf-node 
evaluator of the search tree: they use a combination of an estimated value function and an estimated policy. 
The latest observations at the root node are then used to update the value and policy functions (leaf-node evaluator) for the next iteration.  
They also give a finite sample analysis. However, their result and ours are quite different: in their analysis\blue{,} the sample complexity of MCTS, 
as well as the approximation power of value/policy architectures, are \emph{imposed as an assumption}; here we \emph{prove} an explicit 
finite-sample bound for MCTS and characterize the non-asymptotic error prorogation under MCTS with non-parametric regression for 
leaf-node evaluation. Therefore, they {\em do not} establish ``strong policy improvement'' property of the MCTS.

Two other closely related papers are \cite{teraoka2014efficient} and \cite{kaufmann2017monte}, which study a simplified MCTS for two-player 
zero-sum games. \blue{There,} the goal is to identify the best action of the root in a \emph{given} game tree. For each leaf node, a stochastic oracle 
is provided to generate i.i.d.\ samples for the \emph{true} reward. \cite{teraoka2014efficient} give a high probability bound on the number of 
oracle calls needed for obtaining $\varepsilon$-accurate score at the root. The more recent paper \cite{kaufmann2017monte} develops refined, 
instance-dependent sample complexity bounds. Compared to classical MCTS (e.g., UCT), both the setting and \blue{the} algorithms in the above papers 
are simpler: the game tree is given in advance, rather than being built gradually through samples; the algorithm proposed in~\cite{teraoka2014efficient} 
operates on the tree in a bottom-up fashion with uniform sampling at the leaf nodes. As a result, the analysis is significantly simpler and it is unclear 
whether the techniques can be extended to analyze other variants of MCTS.

It is important to mention the work of \cite{chang2005adaptive} that explores the idea of using UCB for adaptive sampling in MDPs. The approximate 
value computed by the algorithm is shown to converge to the optimal value. We remark that their algorithm is different from the algorithm we analyze 
in this paper. In particular, their algorithm proceeds in a depth-first, recursive manner, and hence involves using UCB for a stationary MAB at each node. 
In contrast, the UCT algorithm we study involves non-stationary MABs, hence our analysis is significantly different from theirs. We refer the readers to 
the work by \cite{kocsis2006bandit} and \cite{coulom2006efficient} for further discussion of this difference. Another related work by \cite{kearns2002sparse} 
studies a sparse sampling algorithm for large MDPs. This algorithm is also different from the MCTS family we analyze in this paper. Recently, 
\cite{efroni2018multiple} study multiple-step lookahead policies in reinforcement learning, which can be implemented via MCTS.  

\subsection{Organization}

Section~\ref{sec:setup} describes the setting of Markov Decision Process considered in this work. 
Section~\ref{sec:mcts} describes the Monte Carlo Tree Search algorithm and the main result about its 
non-asymptotic analysis. Section \ref{sec:RL} describes a reinforcement learning method that combines
the Monte Carlo Tree Search with nearest neighbor supervised learning. It describes the finite-sample 
analysis of the method for finding $\varepsilon$ approximate value function with respect to $\ell_\infty$
norm. Section \ref{sec:nonstatmab} \blue{introduces} a form of non-stationary multi-arm bandit and an upper confidence
bound policy for it. For this setting, we present the concentration of induced regret which serves as a key
result for establishing the property of MCTS. The proofs \blue{of all the technical results} are delegated to Sections \ref{sec:proof.mab} {- \ref{sec:proof.rl} and Appendices.}

%% file: preliminary.tex
\section{Setup \blue{and} Problem Statement} \label{sec:setup}

\subsection{Formal Setup}

We consider the setup of discrete-time discounted Markov decision process (MDP). An MDP is described by a five-tuple $(\mS,\mA,\mP,\mR,\gamma)$, where 
$\mS$ is the set of states, $\mA$ is the set of actions, $\mP\equiv\mP (s'|s,a)$ is the Markovian transition kernel, $\mR\equiv\mR(s,a)$ is a random reward 
function, and $\gamma\in (0,1)$ is a discount factor. \blue{At} each time step, the system is in some state $s \in \mS$. When an action $a\in \mA$ is taken, 
the state transits to a next state $s'\in \mS$ according to the transition kernel $\mP$ and an immediate reward is generated according to $\mR(s,a)$. 

A stationary policy $\pi(a|s)$ gives the probability of performing action $a\in \mA$ given the current state $s\in \mS.$ The \textit{value} function for 
each state $s\in \mS$ under policy $\pi$, denoted by $V^{\pi} (s)$, is defined as the expected discounted sum of rewards received following the policy 
$\pi$ from initial state $s$, i.e., 
\begin{align*}
    V^{\pi}(s)=\E_{\pi}\Big[\sum_{t=0}^{\infty} \gamma^t \mR(s_t,a_t) | s_0=s \Big]. 
\end{align*}
The goal is to find an optimal policy $\pi^{*}$ that maximizes the value from each initial state. The optimal value function 
$V^{*}$ is defined as $V^{*}(s)=V^{\pi^{*}}(s)=\sup_{\pi}V^{\pi}(s)$, $\forall s\in\mS.$ It is well understood that such an optimal policy
exists in reasonable generality. In this paper, we restrict our attention to the MDPs with the following assumptions:
\begin{assumption}[MDP Regularity]
\label{assu:MDP-Reularity}
(A1.) The action space $\mA$ is a finite set and \blue{the} state space $\mS$ is a compact subset of $d$ dimensional \blue{set;} without loss of generality\blue{,} let $\mS = [0,1]^d$;
(A2.) The immediate rewards are random \blue{variables,} uniformly bounded such that 
$\mR(s,a)\in[- R_{\max},  R_{\max}],~\forall s\in\mS,a\in\mathcal{A}$ for some $R_{\max} > 0$; 
(A3.) The state transitions are deterministic, i.e. $\mP\equiv\mP (s'|s,a) \in \{0,1\}$ for all $s, s' \in \mS, ~a \in \mA$.
\end{assumption}
Define 
$\beta \triangleq 1/(1-\gamma) $ and
$V_{\max} \triangleq \beta R_{\max}.		
$ 
Since all the rewards are bounded by $R_{\max}$, it is easy to see that the absolute value of \blue{the} value function for any state under any policy is bounded by $V_{\max}$~\citep{Even-Dar2004Qlearning,Strehl2006Pac}. 

\medskip
\noindent{\bf On Deterministic Transition.}
{We note that
the deterministic transition in MDP should not be viewed as restriction or assumption. Traditional AI game research has been focused on deterministic games with a tree representation. It is this context within which historically MCTS was introduced, 
has been extensively studied and utilized in practice~\citep{browne2012survey}. This includes the recent successes of MCTS in Go~\citep{silver2017mastering}, Chess~\citep{silver2017chess} and Atari games~\citep{guo2014deep}. 
There is a long theoretical literature on the analysis of MCTS and related methods \citep{browne2012survey,hren2008optimistic,munos2014bandits,bartlett2019scale} 
that considers deterministic transition. The principled extension of MCTS algorithm itself as well as theoretical results similar to ours for the stochastic setting are important future work.}

\subsection{Value Function Iteration}

A classical approach to find optimal value function, $V^*$, is an iterative approach called value function iteration. The Bellman equation
characterizes the optimal value function as
\begin{align}\label{eq:bellman}
V^*(s) & = \max_{a \in \mA} \Big( \mathbb{E}[\mR(s, a)] + \gamma V^*(s \circ a) \Big),
\end{align} 
where $s \circ a \in \mS$ is the notation to denote the state reached by applying action $a$ on state $s$. Under Assumption 
\ref{assu:MDP-Reularity}, the transitions are deterministic and hence $s\circ a$ represents a single, deterministic state rather than a random state. 

The value function iteration effectively views  \eqref{eq:bellman} as a fixed point equation and tries to find a solution to it through a natural iteration. 
Precisely, let $V^{(t)}(\cdot)$ be the value function estimation in iteration $t$ \blue{with $V^{(0)}$ 
being arbitrarily initialized}. 
Then, 
for $t \geq 0$, for all $s \in \mS$, 
\begin{align}\label{eq:value.iter_v1}
V^{(t+1)}(s) & = \max_{a \in \mA} \Big(\mathbb{E}[\mR(s, a)] + \gamma V^{(t)}(s \circ a)\Big).
\end{align}
It is well known (cf. \cite{bertsekas2017dynamic}) that value iteration is contractive with respect to $\|\cdot\|_\infty$ norm for all $\gamma < 1$. Specifically, 
for $t \geq 0$, \blue{we have}
\begin{align}\label{eq:value.iter.1_v1}
\| V^{(t+1)} - V^*\|_\infty & \leq \gamma \| V^{(t)} - V^*\|_\infty.
\end{align}

%% file: mctsalgo.tex
\section{Monte Carlo Tree Search} \label{sec:mcts}

Monte Carlo Tree Search (MCTS) has been quite popular recently in many of the reinforcement learning tasks. In effect, given a state $s \in \mS$ and a 
value function estimate $\hat{V}$, it attempts to run the value function iteration for a fixed number of steps, say $H$, to evaluate $V^{(H)}(s)$ starting
with $V^{(0)} = \hat{V}$ per \eqref{eq:value.iter_v1}. This, according to \eqref{eq:value.iter.1_v1}, would provide an estimate within \blue{error} $\gamma^H \|\hat{V} - V^*\|_\infty$ 
-- an excellent estimate of $V^*(s)$ if $H$ is large enough. The goal is to perform computation for value function iteration necessary to
evaluate $V^{(H)}$ for state $s$ only and not necessarily for all states as required by traditional value function iteration. MCTS achieves this by simply 
`unrolling' the associated `computation tree'. Another challenge that MCTS overcomes is the fact that value function iteration as in \eqref{eq:value.iter_v1} 
assumes knowledge of model so that it can compute $\mathbb{E}[\mR(\cdot, \cdot)]$ for any state-action pair. But in reality, rewards are observed through
samples, not a direct access to $\mathbb{E}[\mR(\cdot, \cdot)]$. MCTS tries to utilize the samples in a careful manner to obtain accurate estimation for $V^{(H)}(s)$ over the
computation tree suggested by the value function iteration as discussed above. The concern of careful use of samples naturally connects it to 
multi-arm bandit like setting. 

Next, we present a detailed description of the MCTS algorithm in Section \ref{ssec:mcts.algo}. This can be viewed as a {\em correction} of the algorithm 
presented in \cite{kocsis2006bandit,kocsis2006improved}. We state its theoretical property in Section \ref{ssec:mcts.analysis}. 

\subsection{Algorithm}\label{ssec:mcts.algo}

\begin{figure*}[htp]
  \centering
    \includegraphics[width=.95\textwidth]{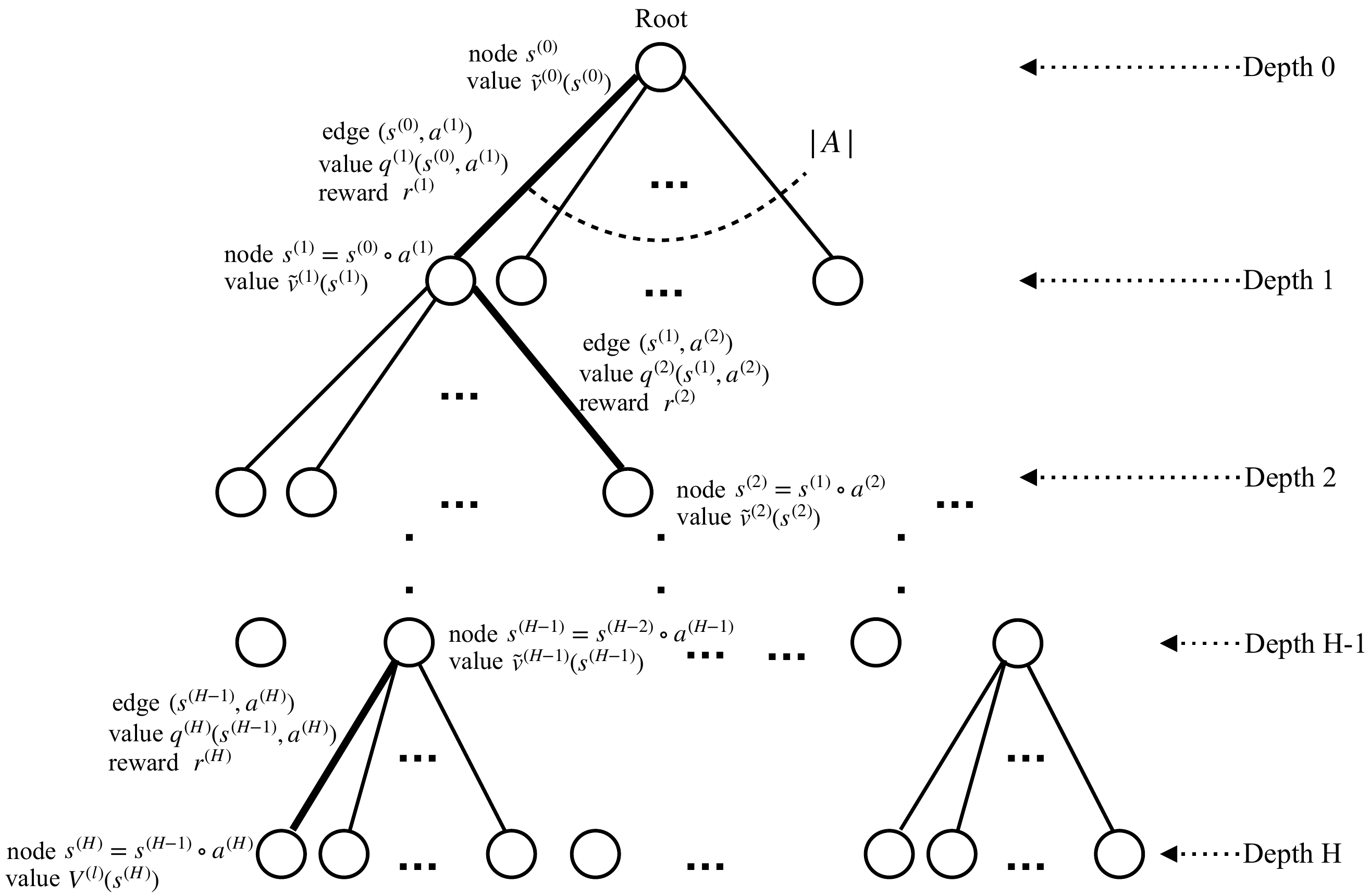}
    \caption{Notation and a sample simulation path of MCTS (thick lines).} \label{fig:setup_mcts}
\end{figure*}

We provide details of a specific form of MCTS, which replaces the logarithmic bonus term of UCT with a polynomial 
one. Overall, we fix the search tree to be of depth $H$. Similar to most literature on this topic, it uses a variant of the Upper Confidence \blue{Bound} (UCB) 
algorithm to select an action at each stage. At a leaf node (i.e., a state at depth $H$), we use the current value oracle \blue{$\hat{V}$} to evaluate its value. Note 
that since we consider deterministic transitions, consequently, the tree is fixed once the root node (state) is chosen, and we use the notation $s\circ a$ 
to denote the next state after taking action $a$ at state $s$. Each edge represents a state-action pair, while each node represents a state. For 
clarity, we use superscript to distinguish quantities related to different depth. The pseudo-code for the MCTS procedure is given in Algorithm \ref{alg:mcts}, and 
Figure \ref{fig:setup_mcts} shows the structure of the search tree and related notation.

\begin{algorithm}[!ht]
   \caption{Fixed-Depth Monte Carlo Tree Search}
   \label{alg:mcts}
\begin{algorithmic}[1]
   \STATE {\bfseries Input:} (1) current value oracle $\hat{V}$, root node $s^{(0)}$ and search depth $H$;\\
   $\quad\quad\quad$(2) number of MCTS simulations $n$;\\
   $\quad\quad\quad$(3) algorithmic constants,
   $\{\alpha^{(i)}\}_{i=1}^H,$ $\{\beta^{(i)}\}_{i=1}^H,$ $\{\xi^{(i)}\}_{i=1}^H$ and $\{\eta^{(i)}\}_{i=1}^H.$
   \STATE {\bfseries Initialization:} for each depth $h$, initialize the cumulative node value $\tilde{v}^{(h)}(s)=0$ and visit count $N^{(h)}(s)=0$ for every node $s$ and initialize the cumulative edge value $q^{(h)}(s,a)=0$. 
   \FOR {each MCTS simulation $t = 1,2 ,\dots, n$}
   \STATE \texttt{/*}~~\texttt{Simulation:}~~\texttt{select actions until reaching depth $H$}\texttt{*/}
   \FOR {depth $h=0,1,2,\dots, H-1$}
   \STATE at state $s^{(h)}$ of depth $h$, select an action (edge) according to
   \begin{equation}
       a^{(h+1)}= \arg\max_{a\in\mathcal{A}}\frac{q^{(h+1)}(s^{(h)},a)+\gamma \tilde{v}^{(h+1)}(s^{(h)}\circ a)}{N^{(h+1)}(s^{(h)}\circ a)}+\frac{\big(\beta^{(h+1)}\big)^{1/\xi^{(h+1)}}\cdot \big(N^{(h)}(s^{(h)})\big)^{\alpha^{(h+1)}/\xi^{(h+1)}}}{\big(N^{(h+1)}(s^{(h)}\circ a)\big)^{1-\eta^{(h+1)}}},\label{eq:alg_mcts}
   \end{equation}
   where dividing by zero is assumed to be $+\infty$.
   \STATE upon taking the action $a^{(h+1)}$, receive a random reward $r^{(h+1)}\triangleq\mathcal{R}(s^{(h)},a^{(h+1)})$ and transit to a new state $s^{(h+1)}$ at depth $h+1$.
   \ENDFOR
  \STATE \texttt{/*}~~\texttt{Evaluation:}~~\texttt{call value oracle for leaf nodes}\texttt{*/}
  \STATE reach $s^{(H)}$ at depth $H$, call the current value oracle and let $\tilde{v}^{(H)}(s^{(H)})=\hat{V}(s^{(H)})$.
   \STATE \texttt{/*}~~\texttt{Update Statistics:}~~\texttt{quantities on the search path}\texttt{*/}
   \FOR {depth $h=0,1,2,\dots, H-1$}
   \STATE update statistics of nodes and edges that are on the search path of current simulation:
   \vspace{-0.08in}
   \begin{align*}
   \vspace{-0.2in}
       \textrm{visit count: }&N^{(h+1)}(s^{(h+1)}) = N^{(h+1)}(s^{(h+1)}) +1\\
       \textrm{edge value: }&q^{(h+1)}(s^{(h)},a^{(h+1)}) = q^{(h+1)}(s^{(h)},a^{(h+1)}) + r^{(h+1)}\\
       \textrm{node value: }&\tilde{v}^{(h)}(s^{(h)})= \tilde{v}^{(h)}(s^{(h)})+ r^{(h+1)} +\gamma r^{(h+2)}+\dots +\gamma^{H-1-h} r^{(H)} +\gamma^{H-h} \tilde{v}^{(H)}(s^{(H)})
       \vspace{-0.2in}
   \end{align*}
   \ENDFOR 
   \ENDFOR
   \STATE {\bfseries Output:} average of the value for the root node ${\tilde{v}^{(0)}(s^{(0)})/}{n}$.
\end{algorithmic}
\end{algorithm}

In Algorithm \ref{alg:mcts}, there are certain sequences of algorithmic parameters required, namely, $\alpha$, $\beta$, $\xi$ and $\eta$. The 
choices for these constants will become clear in our non-asymptotic analysis. At a higher level, the constants for the last layer (i.e., depth $H$), 
$\alpha^{(H)}$, $\beta^{(H)}$, $\xi^{(H)}$ and $\eta^{(H)}$ depend on the properties of the leaf nodes, while the rest are recursively determined 
by the constants one layer below. 
We note that in selecting action $a^{(h+1)}$ at each depth $h$ (i.e., Line 6 of Algorithm \ref{alg:mcts}), the upper confidence term is polynomial in $n$ 
while a typical UCB algorithm would be logarithmic in~$n$, where $n$ is the number of visits to the \blue{corresponding} state thus far. The logarithmic factor in the original UCB algorithm was motivated by the exponential tail 
probability bounds. In our case, it turns out that exponential tail bounds for each layer seems to be infeasible without further structural assumptions.
As mentioned in Section \ref{subsec:related_work}, prior work \citep{audibert2009exploration,salomon2011deviations} has justified the polynomial 
concentration of the regret for the classical UCB in stochastic (independent rewards) multi-arm bandit setting. This implies that the concentration at intermediate 
depth (i.e., depth less than $H$) is at most polynomial. Indeed, we will prove these polynomial concentration bounds even for non-stationary (dependent, non-stationary
rewards) multi-arm
bandit that show up in MCTS and discuss separately in Section \ref{sec:nonstatmab}.  

\subsection{Analysis}\label{ssec:mcts.analysis}
Now, we state the following result on the non-asymptotic performance of the MCTS as described above. 
\begin{thm}\label{thm:MCTS} Consider an MDP satisfying Assumption \ref{assu:MDP-Reularity}. Let $H \geq 1$\blue{, and for $1/2\leq \eta < 1$, let}
\begin{align}
\eta^{(h)} & =\eta^{(H)}\equiv\eta,\quad\forall\: h\in[H], \label{eq:mcts_sequence_1}\\
\alpha^{(h)} & =\eta(1-\eta)\big(\alpha^{(h+1)}-1\big),\quad\forall\: h\in[H-1], \label{eq:mcts_sequence_3}\\
\xi^{(h)} & =\alpha^{(h+1)}-1,\quad\forall h\in[H-1]. \label{eq:mcts_sequence_4}
\end{align}
Suppose that a large enough $\xi^{(H)}$ is chosen such that $\alpha^{(1)}>2$. Then, there exist corresponding constants 
$\{\beta^{(i)}\}_{i=1}^H$ such that for each query state $s\in\mS,$ the following claim holds for the output $\hat{V}_{n}(s)$
of MCTS with $n$ simulations:
\[
	\Big|\E\big[\hat{V}_{n}(s)\big]-V^{*}(s)\Big|\leq\gamma^{H}\varepsilon_{0}+O\Big(n^{\eta-1}\Big),
\]
where $\varepsilon_{0} = \|\hat{V} - V^*\|_\infty$ with $\hat{V}$ being the estimate of $V^*$ utilized by \blue{the} MCTS algorithm
for leaf nodes. 
\end{thm}

Since $\eta \in [1/2, 1)$, Theorem \ref{thm:MCTS} implies a best case convergence rate of $O(n^{-1/2})$ by setting $\eta=1/2$. With these parameter choices, the bias term in the upper confidence bound (line 6 of Algorithm~\ref{alg:mcts}) scales as $\big(N^{(h)}(s^{(h)})\big)^{1/4}/\sqrt{N^{(h+1)}(s^{(h)}\circ a)}$, that is, in the form of $t^{1/4}/\sqrt{S}$ as mentioned in the introduction, where $t\equiv N^{(h)}(s^{(h)})$ is the number of times that state $s^{(h)}$ at depth $h$ has been visited, and $S \equiv N^{(h+1)}(s^{(h)}\circ a)$ is the number of times action $a$ has been selected at state $s^{(h)}$. 

%% file: rlalgo.tex
\section{Reinforcement Learning through MCTS with Supervised Learning} \label{sec:RL}

Recently, MCTS has been utilized prominently in various empirical successes of reinforcement learning including 
AlphaGo Zero (AGZ). Here, MCTS is combined with expressive supervised learning method to iteratively 
improve the policy as well as \blue{the} value function estimation. In effect, MCTS combined with supervised learning 
acts as a ``policy improvement'' operator. 

Intuitively, MCTS produces an improved estimation of value function for a given state of interest\blue{, starting} with a given estimation of value function by ``unrolling'' the ``computation tree'' associated with value function iteration. And MCTS achieves this using observations obtained through simulations. Establishing this \blue{improvement property} rigorously was the primary goal of Section \ref{sec:mcts}. 
\blue{Now,} given such improved estimation of value function for finitely many states, a good supervised learning method can \blue{learn to}
generalize such an improvement to all states. If so, this is like performing value function iteration, but using simulation\blue{s}. 
Presenting such a policy and establishing such guarantee\blue{s} is the crux of this section. 

To that end, we present a reinforcement learning method that combines MCTS with nearest neighbor supervised learning. 
For this method, we establish that indeed\blue{, with} sufficient number of samples, the resulting policy improves the value 
function estimation just like value function iteration. Using this, we provide a finite\blue{-sample} analysis for learning the 
optimal value function within a given tolerance. We find it nearly matching a minimax lower bound in \cite{shah2018qlnn} which we recall in Section \ref{ssec:lb}\blue{, and} thus establishes near minimax optimality of such a reinforcement learning method. 

\subsection{Reinforcement Learning Policy}\label{ssec:rl.policy}
 
Here we describe the policy to produce estimation of optimal value function $V^*$. Similar approach can be applied to obtain estimation of
policy as well. \blue{Let} $V^{(0)}$ be \blue{the} initial estimation of $V^*$, {and for simplicity, let} $V^{(0)}(\cdot) = 0$. We describe a policy that
iterates between use of MCTS and supervised learning to iteratively obtain estimation $V^{(\ell)}$ for $\ell \geq 1$, so that iteratively better estimation of $V^*$ is produced as $\ell$ increases. To that end, for $\ell \geq 1$, 
\begin{itemize}

  \item[$\circ$] For appropriately sampled states $S^\ell = \{s_i\}_{i=1}^{m_{\ell}}$, apply MCTS to obtain improved estimations of value function $\{\hat{V}^{(\ell)}(s_i)\}_{i=1}^{m_{\ell}}$
  using $V^{(\ell-1)}$ \blue{to evaluate leaf nodes during simulations}. 
  
  \item[$\circ$] Using $\{(s_i, \hat{V}^{(\ell)}(s_i)\}_{i=1}^{m_{\ell}}$ with a variant of nearest neighbor supervised learning with parameter $\delta_\ell \in (0,1)$, 
  produce estimation $V^{(\ell)}$ of the optimal value function.
  
\end{itemize}

{For completeness, the pseudo-code is provided in Algorithm \ref{alg:bsrl}.} 
\begin{algorithm}[htp]
   \caption{Reinforcement Learning Policy}
   \label{alg:bsrl}
\begin{algorithmic}[1]
   \STATE {\bfseries Input:} initial value function oracle $V^{(0)}(s)= 0$, $\forall\: s\in\mathcal{S}$ 
   \FOR {$l = 1,2 ,\dots, L$}
   \STATE \texttt{/*}~~\texttt{improvement via MCTS}~~\texttt{*/}
   \STATE uniformly and independently sample states $S^\ell = \{s_i\}_{i=1}^{m_{\ell}}$.
   \FOR {each sampled state $s_i$}
   \STATE apply the MCTS algorithm, which takes as inputs the current value oracle $V^{(l-1)}$, the depth $H^{(l)}$, the number of simulation $n_l$, and the root node $s_i$, and outputs an improved estimate for $V^*(s_i)$:
   \begin{equation}
   \vspace{-0.1in}
       \hat{V}^{(l)}(s_i) = \textrm{MCTS}\big(V^{(l-1)}, H^{(l)}, n_l, s_i\big)
       \vspace{-0.1in}
   \end{equation}
   \ENDFOR
   \STATE \texttt{/*}~~\texttt{supervised learning}~~\texttt{*/}
   \STATE with the collected data $\mathcal{D}^{(l)}=\{(s_i,\hat{V}^{(l)}(s_i))\}_{i=1}^{m_l}$, build a new value oracle $V^{(l)}$ via a nearest neighbor regression with parameter $\delta_l$ :
   \begin{equation}
      \vspace{-0.1in}
       V^{(l)}(s) = \textrm{Nearest Neigbhor}\big(\mathcal{D}^{(l)},\delta_l, s\big)
       ,\: \forall\:s\in\mathcal{S}.
      \vspace{-0.1in}
   \end{equation}
   \ENDFOR
   \STATE {\bfseries Output:} final value oracle $V^{(L)}$.
\end{algorithmic}
\end{algorithm}

\subsection{Supervised Learning}\label{ssec:sl.policy}

For simplicity, we shall utilize the following variant of the nearest neighbor supervised learning parametrized by $\delta \in (0,1)$. Given state space $\mS = [0,1]^d$, we wish to
cover it with minimal (up to scaling) number of balls of radius $\delta$ {(with respect to $\ell_2$-norm)}. To that end, since $\mS = [0,1]^d$, one such construction is where we have balls of radius $\delta$ with
centers being $\{(\theta_1, \theta_2, \dots, \theta_d): \theta_1, \dots, \theta_d \in {\mathcal{Q}}(\delta) \}$ where 
$${\mathcal{Q}}(\delta) = \Big\{\frac{1}{2} \delta i~:~i \in \mathbb{Z}, 0\leq i \leq \Big\lfloor \frac{2}{\delta} \Big\rfloor \Big\} \cup \Big\{1 - \frac{1}{2} \delta i~:~i \in \mathbb{Z}, 0\leq i \leq \Big\lfloor \frac{2}{\delta} \Big\rfloor \Big\}.$$
Let the collection of these balls be denoted by $c_1, \dots, c_{K(\delta, d)}$ with $K(\delta, d) = | \mathcal{Q}(\delta) |$.
It is easy to verify that $\mS \subset \cup_{i\in [K(\delta, d)]} c_i$, $K(\delta, d) = \Theta(\delta^{-d})$ and 
$C_d \delta^{d} \leq {\sf volume}(c_i \cap \mS) \leq C^{\prime}_d \delta^{d}$ for strictly positive constants $C_d, C^{\prime}_d$ that depends on $d$ but not $\delta$.  For any $s \in \mS$, let $j(s) = \min\{j : s \in c_j\}$. 
Given observations $\{(s_i, \hat{V}^{(\ell)}(s_i)\}_{i=1}^{m_{\ell}}$, we produce an estimate $V^{(\ell)}(s)$ for
all $s \in \mS$ as follows: 
\begin{align}
V^{(\ell)}(s) & = \begin{cases} 
				\frac{\sum_{i: s_i \in c_{j(s)}} \hat{V}^{(\ell)}(s_i)}{|\{i: s_i \in c_{j(s)}\}|}, & ~~\mbox{if}~~|\{i: s_i \in c_{j(s)} \}| \neq 0, \\
				0 & ~~\mbox{otherwise.}
			\end{cases} 
\end{align}

\subsection{Finite\blue{-Sample} Analysis}

For finite\blue{-sample} analysis of the proposed reinforcement learning policy, 
we make the following structural assumption about the MDP. Specifically, we 
assume that the optimal value function (i.e., true regression function) is smooth in
some sense.  {We note that some form of smoothness assumption for MDPs with continuous state/action space is typical for $\ell_{\infty}$ guarantee.} The Lipschitz continuous assumption stated below is natural and representative 
in the literature on MDPs with continuous state spaces, cf. ~\cite{munos2014bandits,Dufour2012AnalAppl,Dufour2013LP}
and~\cite{Bertsekas1975TACdiscretization}. 
\begin{assumption}[Smoothness] 
\label{assu:smooth}
The optimal value function $V^{*}:\mathcal{S}\rightarrow\mathbb{R}$ satisfies Lipschitz continuity with parameter 
$C$, i.e., $\forall s,s'\in\mathcal{S} = [0,1]^d,$
$
|V^{*}(s)-V^{*}(s')|\leq C\|s-s'\|_2.
$
\end{assumption}
Now we state the result characterizing the performance of the reinforcement learning policy described above. 
\begin{thm}
\label{thm:MCTS_SL_deterministic} Let Assumptions \ref{assu:MDP-Reularity} and \ref{assu:smooth} hold. \blue{Let
$\varepsilon > 0$ be a given error tolerance}. Then, for $L = \Theta\Big(\log\frac{\varepsilon}{V_{\max}}\Big)$, with appropriately
chosen $m_\ell, \delta_\ell$ for $\ell \in [L]$ as well as parameters in MCTS, the reinforcement learning algorithm produces estimation of value
function $V^{(L)}$ such that 
\begin{align*}
   \E\big[\sup_{s\in\mS}|V^{(L)}(s)-V^*(s)|\big] \leq  \varepsilon, 
\end{align*}
by selecting $m_\ell$ states uniformly at random in $\mS$ \blue{within} iteration $\ell$. This, in total, 
requires \blue{T number of state transitions (or samples),} where 
\begin{align*}
T& =  O\Big(\varepsilon^{-\big(4+d\big)}\cdot \big(\log{\frac{1}{\varepsilon}}\big)^5 \Big).
\end{align*}
 
\end{thm}

\subsection{Minimax Lower Bound}\label{ssec:lb}

Leveraging the the minimax lower bound for the problem of non-parametric regression \citep{tsybakov2009nonparm,stone1982optimal}, recent work \cite{shah2018qlnn} establishes a lower bound on the sample complexity for reinforcement learning algorithms {for general MDPs without additional structural assumptions. Indeed the lower bound also holds for MDPs with deterministic transitions {(the proof is provided in Appendix~\ref{appendix:proof_lower_bound})},} which is stated in the following proposition.


\begin{prop} 
\label{prop:lower_bound}
    Given an algorithm, let $V_T$ be the estimation of $V^*$ after $T$ samples of state transitions for the given MDP. Then, for each $\varepsilon\in(0,1)$, there exists an instance of {deterministic} MDP such that in order to achieve 
	$
	\P\big[\big\Vert \hat{V}_T-V^{*}\big\Vert_{\infty}<\varepsilon\big]\ge 1-\varepsilon,
	$
    it must be that
	\[
	T\ge C'd\cdot\varepsilon^{-(d+2)}\cdot\log(\varepsilon^{-1}),
	\]
	where $C'>0$ is a constant independent of the algorithm.
\end{prop}

Proposition \ref{prop:lower_bound} states that
for any policy to learn the optimal value function within $\varepsilon$ approximation error, the number of samples required must scale as $\widetilde{\Omega}\big({\varepsilon^{-(2+d)}}\big)$. Theorem \ref{thm:MCTS_SL_deterministic} implies that the sample complexity of the proposed algorithm scales as $\widetilde{O}\big({\varepsilon^{-(4+d)}}\big)$ (omitting the logarithmic factor). 
Hence, in terms of the dependence on the dimension, the proposed algorithm is nearly optimal. Optimizing the dependence of the sample complexity on other parameters is an important direction for future work.

%% file: nonstatmab.tex
\section{Non-stationary Multi-Arm Bandit}
\label{sec:nonstatmab}

We introduce a class of non-stationary multi-arm bandit (MAB) problems, which will play a crucial role in analyzing the MCTS algorithm. To that end, let there
be $K \geq 1$ arms or actions of interest. Let $X_{i, t}$ denote the random reward obtained by playing the arm $i \in [K]$ for the $t$th time
with $t \geq 1$. Let
$\bar{X}_{i,n}=\frac{1}{n}\sum_{t=1}^{n}X_{i,t}$ denote the empirical average of playing arm $i$ for $n$ times, and let $\mu_{i,n}=\E[\bar{X}
_{i,n}]$ be its expectation. For each arm $i\in[K]$, the reward $X_{i,t}$ is bounded in $[-R,R]$ for some $R > 0$, and we assume that the reward sequence, $\{X_{i,t}: t \geq 1\}$, is a non-stationary process satisfying the following convergence and concentration properties:
\begin{enumerate}
\item[A.] (Convergence)  the expectation
$\mu_{i,n}$ converges to a value $\mu_{i}$, i.e.,
\begin{align}\label{eq:nonstmab.conv}
&\mu_{i} =\lim_{n\rightarrow\infty}\E[\bar{X}_{i,n}].
\end{align}

\item[B.] (Concentration) there exist three constants, $\beta>1$, $\xi>0$, and $1/2\leq\eta<1$ such that for 
every $z\geq 1$ and every integer $n\geq 1$,
\begin{align}\label{eq:nonstmab.conc}
\P\big(n\bar{X}_{i,n}-n\mu_{i} & \geq n^\eta z\big) \leq\frac{\beta}{z^{\xi}}, ~~ 
\P\big(n\bar{X}_{i,n}-n\mu_{i}\leq-n^\eta z\big) \leq\frac{\beta}{z^{\xi}}.
\end{align}
\end{enumerate}

\subsection{Algorithm}

Consider applying the following variant of Upper Confidence Bound (UCB) algorithm to the above non-stationary MAB. Define
upper confidence bound for arm or action $i$ when it is played $s$ times in total of $t\geq s$ time steps as 
\begin{align}\label{eq:ucb}
U_{i, s, t} & = \bar{X}_{i,s}+B_{t,s},
\end{align}
where $B_{t,s}$ is the ``bonus term". Denote by $I_t$ the arm played at time $t \geq 1$. Then,
\begin{equation}
I_{t}\in\arg\max_{i\in[K]}\big\{\bar{X}_{i,T_{i}(t-1)}+B_{t-1,T_{i}(t-1)}\big\},\label{eq:app_non_ucb}
\end{equation}
where $T_i(t)=\sum_{l=1}^t\mathbb{I}{\{I_l = i\}}$ is the number of times arm $i$ has been played, up to (including) time $t$. We shall make
specific selection of the bonus or bias term $B_{t,s}$ as 
\begin{align}\label{eq:bonus}
B_{t,s}=\frac{  \beta^{1/\xi}\cdot t^{\alpha/\xi}}{s^{1-\eta}}.
\end{align}
A tie is broken arbitrarily when selecting an arm. In the above, $\alpha > 0$ is a tuning parameter that controls the exploration and exploitation trade-off. Let $\mu_*=\max_{i\in[K]}\mu_i$ 
be the optimal value with respect to the converged expectation, and $i_*\in\arg\max_{i\in[K]}\mu_i$ be the corresponding optimal arm. We assume that the optimal arm is unique. Let $\delta_{i*,n}=\mu_{i_*,n}-\mu_{i_*}$, which measures how fast the mean of the optimal non-stationary arm converges. {For simplicity, quantities related to the optimal arm $i_*$ will be simply denoted with subscript $*$, e.g., $\delta_*,n=\delta_{i_*,n}$.} Finally, denote by $\Delta_{\min}=\min_{i\in[K], i\neq i_*} \Delta_i$  the gap between the optimal arm and the second optimal arm with notation  $\Delta_i = \mu_*-\mu_i$. 

\subsection{Analysis}

Let $\bar{X}_{n}\triangleq \frac{1}{n}\sum_{i=1}^KT_i(n)\bar{X}_{i,T_i(n)}$ denote the empirical average under the UCB algorithm (\ref{eq:app_non_ucb}). Then, $\bar{X}_n$ satisfies the following convergence and concentration properties.
\begin{thm}\label{thm:non.stat.mab}
Consider a non-stationary MAB satisfying \eqref{eq:nonstmab.conv} and  \eqref{eq:nonstmab.conc}. Suppose that  algorithm (\ref{eq:app_non_ucb})  is applied with 
parameter $\alpha$ such that $\xi\eta(1-\eta)\leq \alpha <\xi(1-\eta)$ and $\alpha >2$. Then, 
the following holds:
\begin{enumerate}
    \item[A.] Convergence:
    \begin{align*}
        \left|\E[\bar{X}_{n}]-\mu_{*}\right|&\leq|\delta_{*,n}|+\frac{2R (K-1)\cdot\Big(\big(\frac{2}{\Delta_{\min}}\cdot \beta^{1/\xi}\big)^{\frac{1}{1-\eta}}\cdot n^{\frac{\alpha}{\xi(1-\eta)}}+\frac{2}{\alpha-2} + 1\Big)}{n}.
    \end{align*}
    \item[B.] Concentration:
    there exist constants, $\beta'>1$ and $\xi'>0$ and $1/2\leq\eta'<1$ such that for every $n\geq 1$ and every $z\geq 1$,
\begin{align*}
\P\big(n\bar{X}_{n}-n\mu_{*} & \geq n^{\eta'} z\big) \leq\frac{\beta'}{z^{\xi'}},
~~~\P\big(n\bar{X}_{n}-n\mu_{*} \leq -n^{\eta'} z\big) \leq\frac{\beta'}{z^{\xi'}},
\end{align*}
where $\eta'=\frac{\alpha}{\xi(1-\eta)}$, $\xi'=\alpha -1$, $\beta'$ depends on 
$R, K, \Delta_{\min}, \beta,$  $\xi, \alpha, \eta$. 
\end{enumerate}
\end{thm}

%% file: proof_nonstatmab.tex

\section{Proof of Theorem \ref{thm:non.stat.mab}}\label{sec:proof.mab}

We establish the convergence and concentration properties of the variant of the Upper Confidence Bound algorithm described
in Section \ref{sec:nonstatmab} and specified through \eqref{eq:ucb}, \eqref{eq:app_non_ucb} and \eqref{eq:bonus}. 

\medskip
\noindent{\bf Establishing the Convergence Property.} We define a useful notation 
\begin{equation}
\Phi(n,\delta)=n^\eta\bigg(\frac{\beta}{\delta}\bigg)^{1/\xi}. \label{eq:Deviation_Poly}
\end{equation}
We begin with a useful lemma, which shows that the probability that a non-optimal arm or action has a large upper confidence is polynomially small. Proof is provided in Section \ref{appx:proof_UCB_Probability}.
\begin{lem}
\label{lem:UCB_Probability} Let $i \in [K], i \neq i_*$ be a sub-optimal arm and define 
\begin{align}\label{eq:A}
A_{i}(t) & \triangleq\min_{u\in\mathbb{N}}\Big\{\frac{\Phi(u,t^{-\alpha})}{u}\leq\frac{\Delta_{i}}{2}\Big\}=\bigg\lceil\Big(\frac{2}{\Delta_{i}}\cdot \beta^{1/\xi}\cdot t^{\alpha/\xi}\Big)^{\frac{1}{1-\eta}}\bigg\rceil.
\end{align}
For each $s$ and $t$ such that, $A_{i}(t)\leq s\leq t$,  we have 
\[
\P(U_{i,s,t}>\mu_{*})\leq t^{-\alpha}.
\]
\end{lem}

\noindent  Lemma \ref{lem:UCB_Probability} implies that as long as each arm is played enough, the sub-optimal ones become less likely to be selected. 
This allows us to upper bound the expected number of sub-optimal plays as follows. 
\begin{lem}
\label{thm:Number_of_play_upper-Poly} Let $i \in [K], i\neq i_{*}$, then 
\[
\E[T_{i}(n)]\leq\Big(\frac{2}{\Delta_{i}}\cdot \beta^{1/\xi}\Big)^{\frac{1}{1-\eta}}\cdot n^{\frac{\alpha}{\xi(1-\eta)}}+\frac{2}{\alpha-2} + 1.
\]
\end{lem}
Proof of Lemma \ref{thm:Number_of_play_upper-Poly} is deferred to Section \ref{appx:proof_Number_of_play}.

\medskip
\noindent{\em Completing Proof of Convergence.} By the triangle inequality,
\[
\left|\mu_{*}-\E[\bar{X}_{n}]\right|=\left|\mu_{*}-\mu_{*,n}\right|+\left|\mu_{*,n}-\E[\bar{X}_{n}]\right|=\left|\delta_{*,n}\right|+\left|\mu_{*,n}-\E[\bar{X}_{n}]\right|.
\]
The second term can be bounded as follows: 
\begin{align}
&n\left|\mu_{*,n}-\E[\bar{X}_{n}]\right| \nonumber\\
& =\Bigg|\E\bigg[\sum_{t=1}^{n}X_{i_*,t}\bigg]-\E\bigg[\sum_{i=1}^{K}T_{i}(n)\bar{X}_{i,T_{i}(n)}\bigg]\Bigg|\nonumber \\
 & \leq\Bigg|\E\bigg[\sum_{t=1}^{n}X_{i_*,t}\bigg]-\E\bigg[T_{*}(n)\bar{X}_{i_*,T_*(n)}\bigg]\Bigg|
+\Bigg|\E\bigg[\sum_{i=1,i\neq i_{*}}^{K}T_{i}(n)\bar{X}_{i,T_{i}(n)}\bigg]\Bigg|\nonumber \\
 & =\Bigg|\E\bigg[\sum_{t=T_{*}(n)+1}^{n}X_{i_*,t}\bigg]\Bigg|+\Bigg|\E\bigg[\sum_{i=1,i\neq i_{*}}^{K}T_{i}(n)\bar{X}_{i,T_{i}(n)}\bigg]\Bigg|. \label{eq:finite_drift_bound-1}
\end{align}
Recall that the reward sequences are assumed to be bounded in $[-R, R]$. Therefore,
the first term of (\ref{eq:finite_drift_bound-1}) can be bounded
as follows:
\[
\Bigg|\E\bigg[\sum_{t=T_{*}(n)+1}^{n}X_{i_*,t}\bigg]\Bigg|\leq\E\bigg[\sum_{t=T_{*}(n)+1}^{n} |X_{i_*,t}|\bigg]\leq R \cdot \E\bigg[\sum_{i=1,i\neq i_{*}}^{K}T_{i}(n)\bigg].
\]
The second term can also be bounded as:
\begin{align*}
\Bigg|\E\bigg[\sum_{i=1,i\neq i_{*}}^{K}T_{i}(n)\bar{X}_{i,T_{i}(n)}\bigg]\Bigg|
&\leq R \cdot\E\bigg[\sum_{i=1,i\neq i_{*}}^{K}T_{i}(n)\bigg].
\end{align*}
Hence, we obtain that
\[
\left|\mu_{*}-\E[\bar{X}_{n}]\right|=\left|\delta_{*,n}\right|+\left|\mu_{*,n}-\E[\bar{X}_{n}]\right|\leq \left|\delta_{*,n}\right| + \frac{2R\cdot\E\big[\sum_{i=1,i\neq i_{*}}^{K}T_{i}(n)\big]}{n}.
\]
Combining the above bounds and Lemma \ref{thm:Number_of_play_upper-Poly} yields the desired convergence result in Theorem \ref{thm:non.stat.mab}.


\medskip
\noindent{\bf Establishing the Concentration Property.} Having proved the convergence property, the next step is to show that a similar concentration 
property (cf. (\ref{eq:nonstmab.conc})) also holds for $\bar{X}_n$. We aim to precisely capture the relationship between the original constants assumed in the assumption and the new 
constants obtained for $\bar{X}_n$.  To begin with, recall the definition of $A_{i}(t)$ in Lemma \ref{lem:UCB_Probability} and define
\begin{align}\label{eq:Amax}
A(t) & =\max_{i\in[K]}A_i(t) =\bigg\lceil\Big(\frac{2}{\Delta_{\min}}\cdot \beta^{1/\xi}\Big)^{\frac{1}{1-\eta}}\cdot t^{\frac{\alpha}{\xi(1-\eta)}}\bigg\rceil.
\end{align}
{It can be checked that replacing $\beta$ with any larger number still makes the concentration inequalities (\ref{eq:nonstmab.conc}) hold.} Without loss of generality, 
we hence let $\beta$ be large enough so that  $\frac{2}{\Delta_{\min}}\cdot \beta^{1/\xi}>1$. We further denote by $N_p$ the first time such that $t\geq A(t)$, i.e.,
\begin{align}\label{eq:Np}
N_p & =\min\{t\geq 1\::\:t\geq A(t)\}~=~\Theta\Big(\big(\frac{2^\xi \beta}{\Delta_{\min}^{\xi}}\big)^{\frac{1}{\xi (1-\eta) - \alpha}}\Big).
\end{align}
We first state the following concentration property, which will be further refined to match the desired form in Theorem \ref{thm:non.stat.mab}. We defer the proof to Section \ref{appx:proof_Concentrate-Poly}.
\begin{lem}\label{lem:Concentrate-Poly}
For any $n\geq N_p$ and $x\geq 1$, let $r_{0}=n^\eta+ 2 R(K-1)\big(3+A(n)\big).$ Then, 
\begin{align*}
& \P\Big(n\bar{X}_{n}-n\mu_*\geq r_0x\Big)\leq\frac{\beta}{x^{\xi}}+\frac{2(K-1)}{(\alpha-1)\big((1+A(n))x\big)^{\alpha-1}},\\
& \P\Big(n\bar{X}_{n}-n\mu_*\leq -r_0x\Big)\leq\frac{\beta}{x^{\xi}}+\frac{2(K-1)}{(\alpha-1)\big((1+A(n))x\big)^{\alpha-1}}.
\end{align*}
\end{lem}

\noindent Lemma \ref{lem:Concentrate-Poly} confirms that indeed, as $n$ becomes large, the average $\bar{X}_n$ also satisfies certain concentration inequalities. 
However, the particular form of concentration in Theorem \ref{lem:Concentrate-Poly} does not quite match the form of concentration in Theorem \ref{thm:non.stat.mab}
which we conclude next. 

\medskip
\noindent{\em Completing Proof of Concentration Property.} Let $N_p'$ be a constant defined as follows:
\begin{equation*}
N_p'=\min\{t\geq 1: t\geq A(t)\textrm{ and } 2R A(t)\geq t^\eta+2R (4K-3)\}.
\end{equation*}
Recall the definition of $A(t)$ and that $\alpha\geq \xi\eta(1-\eta)$ and $\alpha<\xi(1-\eta)$. Hence, $N_p'$ is guaranteed to exist. In addition, note that by definition, $N_p'\geq N_p$. For each $n\geq N_p'$,
\begin{align*}
2R K \Big(\frac{2}{\Delta_{\min}}\cdot \beta^{1/\xi}\Big)^{\frac{1}{1-\eta}}\cdot n^{\frac{\alpha}{\xi(1-\eta)}}&= 2R K \Big[\Big(\frac{2}{\Delta_{\min}}\cdot \beta^{1/\xi}\Big)^{\frac{1}{1-\eta}}\cdot n^{\frac{\alpha}{\xi(1-\eta)}}+1-1\Big]\\
&\geq 2R KA(n)-2R K\\
&= 2R(K-1)A(n)+2R A(n)-2R K\\
&\geq 2R (K-1)A(n) + n^\eta+2R (4K-3)-2R K\\
& = 2R(K-1)(A(n)+3) + n^\eta= r_0
\end{align*}  
Now, let us apply Lemma \ref{lem:Concentrate-Poly}: for every $n\geq N_p'$ and $x\geq 1$, we have 
\begin{align}
  \P\Big(n\bar{X}_{n}-n\mu_*\geq n^{\frac{\alpha}{\xi(1-\eta)}}\Big[2R K \Big(\frac{2}{\Delta_{\min}}\cdot \beta^{1/\xi}\Big)^{\frac{1}{1-\eta}}\Big]x \Big)
&\leq\P\Big(n\bar{X}_{n}-n\mu_*\geq r_0x\Big) \nonumber \\
&\leq  \frac{\beta}{x^{\xi}}+\frac{2(K-1)}{(\alpha-1)\big((1+A(n))x\big)^{\alpha-1}}\nonumber\\
&\leq  \frac{2\max(\beta,\frac{2(K-1)}{(\alpha-1)(1+A(N_p'))^{\alpha-1}})}{x^{\alpha-1}}, \label{eq:theorem_6_proof_1}
\end{align}
where the last inequality follows because $n\geq N_p'$  and $A(n)$ is a non-decreasing function. In addition, since $\alpha< \xi(1-\eta)<\xi$, we have $\alpha-1<\xi$. For convenience, we define a constant
\begin{equation}
c_1\triangleq 2R K \Big(\frac{2}{\Delta_{\min}}\cdot \beta^{1/\xi}\Big)^{\frac{1}{1-\eta}}.\label{eq:thm6_c_1}
\end{equation}
Equivalently, by a change of variable, i.e., letting 
$z=c_1x$,
then for every $n\geq N_p'$ and $z\geq 1$, we obtain that
\begin{align}
  \P\Big(n\bar{X}_{n}-n\mu_*\geq n^{\frac{\alpha}{\xi(1-\eta)}}z \Big)
&\leq  \frac{2c_1^{\alpha-1}\cdot\max\big(\beta,\frac{2(K-1)}{(\alpha-1)(1+A(N_p'))^{\alpha-1}}\big)}{z^{\alpha-1}}. \label{eq:theorem_6_proof_2}
\end{align}
The above inequality holds because: (1) if $z\geq c_1$, then (\ref{eq:theorem_6_proof_2}) directly follows from (\ref{eq:theorem_6_proof_1}); (2) if $1\leq z\leq c_1$, then the R.H.S. of (\ref{eq:theorem_6_proof_2}) is at least 1 (by assumption, $\beta>1$) and the inequality trivially holds.  
The concentration inequality, i.e., Eq.~(\ref{eq:theorem_6_proof_2}), is now almost the same as the desired form in Theorem \ref{thm:non.stat.mab}. The only difference is that it only holds for $n\geq N_p'$. This is not hard to resolve. The easiest approach, which we show in the following, 
is to refine the constants to ensure that when $1\leq n< N_p'$, Eq.~(\ref{eq:theorem_6_proof_2}) is trivially true.
To this end, we note that $|n\bar{X}_n-n\mu_*|\leq 2R n$. For each $1\leq n<N_p'$, there is a corresponding $\bar{z}(n)$ such that $n^{\frac{\alpha}{\xi(1-\eta)}}\bar{z}(n)=2 R n$. That is,
\begin{equation*}
\bar{z}(n)\triangleq 2R n^{1-\frac{\alpha}{\xi(1-\eta)}}, \quad  1\leq n<N_p'.
\end{equation*}
This then implies that for each $1\leq n<N_p'$, the following inequality trivially holds:
\begin{equation*}
\P\Big(n\bar{X}_{n}-n\mu_*\geq n^{\frac{\alpha}{\xi(1-\eta)}}z \Big)
\leq \frac{\bar{z}(n)^{\alpha-1}}{z^{\alpha-1}}, \quad \forall\:z\geq 1.
\end{equation*}
To see why, note that for each $1\leq n<N_p'$: (1) if $z\geq \bar{z}(n)$, then $n^{\frac{\alpha}{\xi(1-\eta)}}z\geq 2R n$ and the above probability should be $0$. Hence, any positive number on the R.H.S. makes the inequality trivially true; (2) if $1\leq z<\bar{z}(n)$, the R.H.S. is at least 1, which again makes the inequality hold. For convenience, define
\begin{equation}
c_2\triangleq \max_{1\leq n< N_p'}\bar{z}(n)=2R (N_p'-1)^{1-\frac{\alpha}{\xi(1-\eta)}}. \label{eq:thm6_c_2}
\end{equation} 
Then, it is easy to see that for every $n\geq 1$ and every $z\geq 1$, we have
\begin{align*}
\P\big(n\bar{X}_{n}-n\mu_{*}\geq n^{\eta'} z\big) \leq\frac{\beta'}{z^{\xi'}},
\end{align*}
where the constants are given by
\begin{align}
\eta' &= \frac{\alpha}{\xi(1-\eta)}, \label{eq:thm6_eta}\\
\xi'&= \alpha -1, \label{eq:thm6_c}\\
\beta'&= \max\Big\{c_2,\:2c_1^{\alpha-1}\cdot\max\big(\beta,\frac{2(K-1)}{(\alpha-1)(1+A(N_p'))^{\alpha-1}}\big)\Big\}.\label{eq:thm6_c0}
\end{align}
Finally, notice that since $\alpha\geq\xi\eta(1-\eta)$ and $\alpha<\xi(1-\eta)$, we have $1/2\leq\eta\leq\eta'<1$. Note that per 
\eqref{eq:thm6_c_1}, $c_1$ depends on $R, K, \Delta_{\min}, \beta, \xi$ and $\eta$. In addition, $c_2$ depends on 
$R, K, \Delta_{\min}, \beta, \xi, \alpha, \eta$ and $N_p'$ depends on $R, K, \Delta_{\min}, \beta, \xi, \alpha, \eta$. Therefore, $\beta'$
depends on $R, K, \Delta_{\min}, \beta,$  $ \xi, \alpha, \eta$.  The other direction follows exactly the same reasoning, and this completes the proof of Theorem \ref{thm:non.stat.mab}.

\subsection{Proof of Lemma ~\ref{lem:UCB_Probability}} \label{appx:proof_UCB_Probability}

By the choice of $A_{i}(t)$, $s$ and $t$, we have $B_{t,s}=\frac{\Phi(s,t^{-\alpha})}{s}\leq\frac{\Phi(A_i(t),t^{-\alpha})}{A_i(t)}\leq\frac{\Delta_{i}}{2}$.
Therefore,
\begin{align*}
\P(U_{i,s,t}>\mu_{*}) & =\P(\bar{X}_{i,s}+B_{t,s}>\mu_{*})\\
 & =\P\bigg(\bar{X}_{i,s}-\mu_{i}>\Delta_{i}-B_{t,s}\bigg)\\
 & \leq\P\bigg(\bar{X}_{i,s}-\mu_{i}>B_{t,s}\bigg) &  & \Delta_{i}\geq2B_{t,s}\\
 & \leq t^{-\alpha}. &  & \text{by concentration (\ref{eq:nonstmab.conc})}.
\end{align*}

\subsection{Proof of Lemma~\ref{thm:Number_of_play_upper-Poly}} \label{appx:proof_Number_of_play}

If a sub-optimal arm $i$ is chosen at time $t+1$, i.e., $I_{t+1}=i$, then at least one of the following two equations must be true: with notation
$T_*(\cdot) = T_{i_*}(\cdot$), 
\begin{align}
U_{i_*,T_{*}(t),t} & \leq\mu_{*} \; , \label{eq:Uopt_bound}\\
U_{i,T_{i}(t),t} & >\mu_{*} \; . \label{eq:Ui_bound}
\end{align}
Indeed, if both inequalities are false, we have $U_{i_*,T_{*}(t),t}>\mu_*\geq U_{i,T_{i}(t),t}$, which is a 
contradiction to $I_{t+1}=i$.  We now use this fact to prove Lemma \ref{thm:Number_of_play_upper-Poly}.

\medskip
\noindent{\bf Case 1:} $n> A_i(n)$. Note that such $n$ exists because $A_i(n)$ grows with a polynomial order $O\big(n^{\frac{\alpha}{\xi(1-\eta)}}\big)$ and $\alpha<\xi(1-\eta)$, i.e., $A_i(n)=o(n)$.  Then,
\begin{align*}
T_{i}(n) & =\sum_{t=0}^{n-1}\indic\{I_{t+1}=i\}\stackrel{(a)}{=}1+\sum_{t=K}^{n-1}\indic\{I_{t+1}=i\}\\
 & {=}1+\sum_{t=K}^{n-1}\big(\indic\{I_{t+1}=i,T_{i}(t)<A_{i}(n)\}+\indic\{I_{t+1}=i,T_{i}(t)\geq A_{i}(n)\}\big)\\
 & \leq A_{i}(n)+\sum_{t=K}^{n-1}\indic\{I_{t+1}=i,T_{i}(t)\geq A_{i}(n)\},
\end{align*}
where equality (a) follows from the fact that $ B_{t,s}=\infty$ if $s=0$.

To analyze the above summation, we note that from (\ref{eq:Uopt_bound}) and (\ref{eq:Ui_bound}),
\begin{align*}
 \indic\{I_{t+1}=i,T_{i}(t)\geq A_{i}(n)\} 
&\leq\indic\{U_{i_*,T_{*}(t),t}\leq\mu_{*}\text{ or }U_{i,T_{i}(t),t}>\mu_{*},T_{i}(t)\geq A_{i}(n)\}\\
 & \leq\indic\{U_{i,T_{i}(t),t}>\mu_{*},T_{i}(t)\geq A_{i}(n)\}+\indic\{U_{i_*,T_{*}(t),t}\leq\mu_{*},T_i(t)\geq A_i(n)\}\\
 & \leq\indic\{U_{i,T_{i}(t),t}>\mu_{*},T_{i}(t)\geq A_{i}(n)\}+\indic\{U_{i_*,T_{*}(t),t}\leq\mu_{*}\}\\
 & =\indic\{\exists\: s:A_{i}(n)\leq s\leq t,\text{ s.t. }U_{i,s,t}>\mu_{*}\} +\indic\{\exists\: s_{*}:1\leq s_{*}\leq t,\text{ s.t. }U_{i_*,s_{*},t}\leq\mu_{*}\}.
\end{align*}

\medskip{}

To summarize, we have proved that
\begin{align}
\E[T_{i}(n)] 
& \leq A_{i}(n)+\sum_{t=A_{i}(n)}^{n-1}\P\Big((\ref{eq:Uopt_bound})\text{ or }(\ref{eq:Ui_bound})\text{ is true}\text{, and }T_{i}(t)\geq A_{i}(n)\Big)\nonumber\\
 & \leq A_{i}(n)+\sum_{t=A_{i}(n)}^{n-1}\Big[\P\big(\underbrace{\exists \:s:A_{i}(n)\leq s\leq t,\text{ s.t. }U_{i,s,t}>\mu_{*}}_{E_{1}}\big)+\P\big(\underbrace{\exists \:s_{*}:1\leq s_{*}\leq t,\text{ s.t. }U_{i_*,s_{*},t}\leq\mu_{*}}_{E_{2}}\big)\Big]. \label{eqn:lemma3_proof_1}
\end{align}
To complete the proof of Lemma \ref{thm:Number_of_play_upper-Poly}, it suffices to bound the probabilities of the two events $E_{1}$
and $E_{2}$. To this end, we use a union bound: 
\begin{align*}
\P\big(E_{1}\big) & \leq\sum_{s=A_{i}(n)}^{t}\P\big(U_{i,s,t}>\mu_{*}\big)
  \overset{(a)}{\le} \sum_{s=A_{i}(n)}^{t}t^{-\alpha} 
  \leq t \cdot t^{-\alpha}
  = t^{1-\alpha},
\end{align*}
where the step $(a)$ follows from $A_i(n)\geq A_i(t)$ and Lemma~\ref{lem:UCB_Probability}.
We bound $\P(E_{2})$ in a similar way: 
\begin{align*}
\P(E_{2}) & \leq\sum_{s_*=1}^{t}\P(U_{i_*,s_*,t}\leq\mu_{*})
 =\sum_{s_*=1}^{t}\P\bigg(\bar{X}_{i_*,s_*}+B_{t,s_*}\leq\mu_{*}\bigg)\overset{(a)}{\leq} \sum_{s_*=1}^{t}t^{-\alpha} 
  \leq t^{1-\alpha},
\end{align*}
where step $(a)$ follows from concentration (cf.\ \eqref{eq:nonstmab.conc}). By substituting the bounds of 
$\P(E_{1})$ and $\P(E_{2})$ into (\ref{eqn:lemma3_proof_1}), we have:
\begin{align*}
\E[T_{i}(n)] & \leq A_{i}(n)+\sum_{t=A_{i}(n)}^{n-1}2t^{1-\alpha}\\
 & \leq A_{i}(n)+\int_{A_{i}(n)-1}^{\infty}2t^{1-\alpha}dt & & \alpha>2\\
 & =A_{i}(n)+\frac{2\big(A_{i}(n)-1\big)^{2-\alpha}}{\alpha-2}\\
 & \leq A_{i}(n)+\frac{2}{\alpha-2}\\
 &\leq \Big(\frac{2}{\Delta_{i}}\cdot \beta^{1/\xi}\Big)^{\frac{1}{1-\eta}}\cdot n^{\frac{\alpha}{\xi(1-\eta)}}+\frac{2}{\alpha-2} + 1.
\end{align*}
\medskip
\noindent{\bf Case 2:} $n\leq A_i(n)$. Note that if $n$ is such that $n\leq A_i(n)$, then the above bound trivially holds because $T_i(n)\leq n\leq A_i(n)$. This completes the proof of Lemma \ref{thm:Number_of_play_upper-Poly}.

\subsection{Proof of Lemma~\ref{lem:Concentrate-Poly}} \label{appx:proof_Concentrate-Poly}

We first prove one direction, namely, $\P(n\mu_*-n\bar{X}_n\geq r_0x)$. The other direction follows the similar steps, and we will comment on that at the end of this proof.
The general idea underlying the proof is to rewrite the quantity $n\mu_*-n\bar{X}_n$ as sums of terms that can be bounded using previous lemmas or assumptions. To begin with, note that
\begin{align*}
n\mu_{*}-n\bar{X}_{n} & =n\mu_{*}-\sum_{i=1}^{K}T_{i}(n)\bar{X}_{i,T_{i}(n)}\\
 & =n\mu_{*}-\sum_{t=1}^{T_{*}(n)}X_{i_*,t}-\sum_{i\neq i_{*}}T_{i}(n)\bar{X}_{i,T_{i}(n)}\\
 & =n\mu_{*}-\sum_{t=1}^{n}X_{i_*,t}+\sum_{t=T_{*}(n)+1}^{n}X_{i_*,t}-\sum_{i\neq i_{*}}\sum_{t=1}^{T_{i}(n)}X_{i,t}\\
 & \leq n\mu_{*}-\sum_{t=1}^{n}X_{i_*,t}+2 R \sum_{i\neq i_{*}}T_{i}(n), 
\end{align*}
because $X_{i, t} \in [-R, R]$ for all $i, t$. Therefore, we have
\begin{align}
\P\big(n\mu_{*}-n\bar{X}_{n}\geq r_{0}x\big) & \leq\P\big(n\mu_{*}-\sum_{t=1}^{n}X_{i_*,t}+2 R \sum_{i\neq i_{*}}T_{i}(n)\geq r_{0}x\big)\nonumber \\
 & \leq\P\big(n\mu_{*}-\sum_{t=1}^{n}X_{i_*,t}\geq n^\eta x\big)
 +\sum_{i\neq i_{*}}\P\big(T_{i}(n)\geq{(3+A(n))x}\big),\label{eq:Concentrate_upb}
\end{align}
where the last inequality follows from the union bound. 

To prove the theorem, we now bound the two terms in (\ref{eq:Concentrate_upb}).
By our concentration assumption, we can upper bound the
first term as follows:
\begin{equation}
\P\big(n\mu_{*}-\sum_{t=1}^{n}X_{i_*,t}\geq n^\eta x\big)\leq\frac{\beta}{x^{\xi}}.\label{eq:Concentrate_opt}
\end{equation}

Next, we bound each term in the summation of (\ref{eq:Concentrate_upb}). Fix $n$ and a sub-optimal edge
$i$. Let $u$ be an integer satisfying $u\geq A(n).$ For any
$\tau\in\mathbb{R}$, consider the following two events: 
\begin{align*}
E_{1} & =\big\{\text{For each integer } t\in [u,n],\text{ we have }U_{i,u,t}\leq\tau\big\},\\
E_{2} & =\big\{\text{For each integer } s\in [1,n-u],\text{ we have }U_{i_*,s,u+s}>\tau\big\}.
\end{align*}
As a first step, we want to show that
\begin{equation}
E_1\cap E_2 \Rightarrow T_i(n)\leq u. \label{eq:thm7_proof_1}
\end{equation}
To this end, let us condition on both events $E_{1}$ and $E_{2}$. Recall that $B_{t,s}$ is
non-decreasing with respect to $t$. Then, 
for each $s$ such that $1\leq s\leq n-u,$
and each $t$ such that $u+s\leq t\leq n$, it holds that 
\[
U_{i_*,s,t}=\bar{X}_{i_*,s}+B_{t,s}\geq\bar{X}_{i_*,s}+B_{u+s,s}=U_{i_*,s,u+s}>\tau\geq U_{i,u,t}.
\]

This implies that $T_i(n)\leq u$. To see why, suppose that $T_i(n) > u$ and denote by $t'$ the first time that arm $i$ has been played $u$ times, i.e., $t'=\min\{t: t \leq n ,T_i(t)=u\}$. Note that by definition $t'\geq u+T_*(t')$. Hence, for any time t such that $t'< t\leq n$, the above inequality implies that $U_{i_*,T_*(t),t}> U_{i,u,t}$. That is, $i^*$ always has a higher upper confidence bound than $i$, and
arm $i$ will not be selected, i.e., arm $i$ will not be played the
$(u+1)$-th time. This contradicts our assumption that $T_i(n)>u$, and hence we have the inequality $T_i(n)\leq u$. 

To summarize, we have established the fact that $E_1\cap E_2 \Rightarrow T_i(n)\leq u.$
As a result, we have:
\begin{align*}
\{T_{i}(n)  >u\}&\subset\big(E_{1}^{c}\cup E_{2}^{c}\big)
 =\big(\big\{\exists\: t:\ u\leq t\leq n\text{ s.t. }U_{i,u,t}>\tau\big\}
 \cup\big\{\exists \:s:\text{}1\leq s\leq n-u,\text{ s.t. }U_{i_*,s,u+s}\leq\tau\big\}\big).
\end{align*}
Using union bound, we obtain that
\begin{equation}
\P\big(T_{i}(n)>u\big)\leq\sum_{t=u}^{n}\P(U_{i,u,t}>\tau)+\sum_{s=1}^{n-u}\P\big(U_{i_*,s,u+s}\leq\tau\big).\label{eq:PTi_bound-1}
\end{equation}
Note that for the above bound, we are free to choose $u$ and $\tau$ as long as $u\geq A(n)$. To connect with our goal (cf.\ (\ref{eq:Concentrate_upb})), in the following, we set $u=\lfloor(1+A(n))x\rfloor+1$ (recall that $x\geq1$) and $\tau=\mu_{*}$ to bound $\P(T_{i}(n)>u).$
Since $u\geq A(n)\geq A_i(n)$, by Lemma \ref{lem:UCB_Probability},
we have 
\begin{align*}
\sum_{t=u}^{n}\P(U_{i,u,t}>\mu_{*})  \leq\sum_{t=u}^{n}t^{-\alpha}\leq\int_{u-1}^{\infty}t^{-\alpha}dt&=\frac{(u-1)^{1-\alpha}}{\alpha-1}\\
& = \frac{(\lfloor(1+A(n))x\rfloor)^{1-\alpha}}{\alpha-1}
\leq \frac{\Big((1+A(n))x\Big)^{1-\alpha}}{\alpha-1}.
\end{align*}
As for the second summation in the R.H.S. of (\ref{eq:PTi_bound-1}), we have that 
\begin{align*}
\sum_{s=1}^{n-u}\P\big(U_{i_*,s,u+s}\leq\tau\big) & =\sum_{s=1}^{n-u}\P(U_{i_*,s,u+s}\leq\mu_{*}) \\
 & = \sum_{s=1}^{n-u}\P\bigg(\bar{X}_{i_*,s}+B_{u+s,s}\leq\mu_{*}\bigg)\\ 
 & \leq\sum_{s=1}^{n-u}(s+u)^{-\alpha} 
 = \sum_{t=1+u}^nt^{-\alpha}\\
 & \leq\int_{u-1}^{\infty}t^{-\alpha}dt=\frac{(u-1)^{1-\alpha}}{\alpha-1}\leq\frac{\Big((1+A(n))x\Big)^{1-\alpha}}{\alpha-1},
\end{align*}
where the first inequality follows from the concentration property, cf. \eqref{eq:nonstmab.conc}.
Combining the above inequalities and note that $(3+A(n))x > \lfloor(1+A(n))x\rfloor+1$:
\begin{equation}
\P\big(T_{i}(n)\geq(3+A(n))x\big)\leq\P\big(T_{i}(n)>u\big)\leq\frac{2\Big(\big(1+A(n)\big)x\Big)^{1-\alpha}}{\alpha-1}.\label{eq:Concentrate_subopt}
\end{equation}
Substituting (\ref{eq:Concentrate_opt}) and (\ref{eq:Concentrate_subopt})
into (\ref{eq:Concentrate_upb}), we obtain
\[
\P\big(n\mu_{*}-n\bar{X}_{n}\geq r_{0}x\big)\leq\frac{\beta}{x^{\xi}}+\sum_{i\neq i_{*}}\frac{2\Big(\big(1+A(n)\big)x\Big)^{1-\alpha}}{\alpha-1},
\]
which is the desired inequality in Lemma \ref{lem:Concentrate-Poly}.

To complete the proof, we need to consider the other direction, i.e., $\P(n\bar{X}_n-n\mu_*\geq r_0x)$. The proof is almost identical. Note that
\begin{align*}
n\bar{X}_{n}-n\mu_* & =\sum_{i=1}^{K}T_{i}(n)\bar{X}_{i,T_{i}(n)}-n\mu_*\\
 & =\sum_{t=1}^{n}X_{i_*,t}-n\mu_*-\sum_{t=T_{*}(n)+1}^{n}X_{i_*,t}+\sum_{i\neq i_{*}}\sum_{t=1}^{T_{i}(n)}X_{i,t}\\
 & \leq \sum_{t=1}^{n}X_{i_*,t}-n\mu_*+2 R \sum_{i\neq i_{*}}T_{i}(n), 
\end{align*}
because $X_{i,t}\in [-R, R]$ for all $i, t$. Therefore, 
\begin{align*}
\P\big(n\bar{X}_{n}-n\mu_*\geq r_{0}x\big) 
&\leq\P\big(\sum_{t=1}^{n}X_{i_*,t}-n\mu_*+ 2 R \sum_{i\neq i_{*}}T_{i}(n)\geq r_{0}x\big)\nonumber \\
 & \leq\P\big(\sum_{t=1}^{n}X_{i_*,t}-n\mu_*\geq n^\eta x\big)+\sum_{i\neq i_{*}}\P\big(T_{i}(n)\geq(3+A_{i}(n))x\big).
\end{align*}
The desired inequality then follows exactly from the same reasoning of our previous proof.

%% file: proof_mcts.tex
\section{Analysis of MCTS and Proof of Theorem~\ref{thm:MCTS}}
\label{sec:proof.mcts}

In this section, we give a complete analysis for the fixed-depth Monte Carlo Tree Search (MCTS) algorithm illustrated in Algorithm \ref{alg:mcts} and prove Theorem \ref{thm:MCTS}. In effect, as discussed in Section \ref{sec:mcts}, one can view a depth-$H$ MCTS as a simulated version of $H$ steps value function iterations. Given the current value function proxy $\hat{V}$, let $V^{(H)}(\cdot)$ be the value function estimation after $H$ steps of value function iteration starting with the proxy 
$\hat{V}$.
Then, we prove the result in two parts. 
 First, we argue that due to the MCTS sampling process, the mean of the empirical estimation of value function at the query node $s$, or the root node of MCTS tree, is within $O(n^{\eta - 1})$ of $V^{(H)}(s)$ after $n$ simulations, with the given proxy 
$\hat{V}$ being the input to the MCTS algorithm. Second, we argue that $V^{(H)}(s)$ is within $\gamma^H \| \hat{V} - V^*\|_\infty \leq \gamma^H \varepsilon_0$ of the optimal value function. Putting this together leads to Theorem \ref{thm:MCTS}. 



We start by a preliminary probabilistic lemma in Section \ref{sec:prelim} that will be useful throughout. 
Sections \ref{ssec:part1.1} and \ref{ssec:part1.2} argue the first part of the proof as explained above. Section \ref{ssec:part2} provides
proof of the second part. And Section \ref{ssec:conc} concludes the proof of Theorem \ref{thm:MCTS}. 

\subsection{Preliminary}\label{sec:prelim}

We state the following probabilistic lemma that is useful throughout. Proof can be found in Section \ref{appx:proof_prelim}.
\begin{lem}\label{lem.prelim}
Consider real-valued random variables $X_i, Y_i$ for $i \geq 1$ such that $X$s are independent and identically distributed taking values in $[-B, B]$ for some $B > 0$,   
$X$s are independent of $Y$s, and $Y$s satisfy 
\begin{enumerate}
    \item[A.] Convergence: for $n \geq 1$, with notation $\bar{Y}_n = \frac1n \big(\sum_{i=1}^n Y_i\big)$, 
    \begin{align*}
        \lim_{n\to\infty} \E[\bar{Y}_{n}] & = \mu_Y.
    \end{align*}
    \item[B.] Concentration: there exist constants, $\beta>1$, $\xi>0$, $1/2\leq \eta<1$ such that for $n\geq 1$ and $z\geq 1$,
\begin{align*}
&\P\big(n\bar{Y}_{n}-n\mu_{Y}\geq n^{\eta} z\big) \leq\frac{\beta}{z^{\xi}}, \quad
\P\big(n\bar{Y}_{n}-n\mu_{Y} \leq -n^{\eta} z\big) \leq\frac{\beta}{z^{\xi}}.
\end{align*}
\end{enumerate}
Let $Z_i = X_i + \rho Y_i$ for some $\rho > 0$. Then, $Z$s satisfy
\begin{enumerate}
    \item[A.] Convergence: for $n \geq 1$, with notation $\bar{Z}_n = \frac1n \big(\sum_{i=1}^n Z_i\big)$, and $\mu_X = \mathbb{E}[X_1]$, 
    \begin{align*}
        \lim_{n\to\infty} \E[\bar{Z}_{n}] & = \mu_X + \rho \mu_Y. 
    \end{align*}
    \item[B.] Concentration: there exist constant $\beta'>1$ depending upon $\rho, \xi, \beta$ and $B$, such that for $n\geq 1$ and $z\geq 1$,
\begin{align*}
&\P\big(n\bar{Z}_{n}-n(\mu_{X} + \rho \mu_{Y}) \geq n^{\eta} z\big) \leq\frac{\beta'}{z^{\xi}}, \quad 
\P\big(n\bar{Z}_{n}-n(\mu_{X} + \rho  \mu_{Y})  \leq -n^{\eta} z\big) \leq\frac{\beta'}{z^{\xi}}.
\end{align*}
\end{enumerate}
\end{lem}

\subsection{Analyzing Leaf Level $H$}\label{ssec:part1.1}

The goal is to understand the empirical reward observed at the query node for MCTS or the root node of the MCTS tree. In particular, 
we argue that the mean of the empirical reward at the root node of the MCTS tree is within $O(n^{\eta -1})$ of the mean
reward obtained at it assuming access to infinitely many samples. We start by analyzing the reward collected
at the nodes that are at leaf level $H$ and level $H-1$. 

The nodes at leaf level, i.e., level $H$, are children of nodes at level $H-1$ in the MCTS tree. 
Let there be $n_{H-1}$ nodes at level $H-1$ corresponding to states $s_{1, H-1},\dots, s_{n_{H-1}, H-1} \in \mS$. Consider
node $i \in [n_{H-1}]$ at level $H-1$, corresponding to state $s_{i, H-1}$. As part of the algorithm, whenever this node is visited, 
one of the $K$ feasible actions is taken. When an action $a \in [K]$ is taken, the node $s'_{H} = s_{i, H-1} \circ a$, at the leaf level $H$ 
is reached. This results in reward at node $s_{i, H-1}$ (at level $H-1$) being equal to $\mR(s_{i, H-1}, a) + \gamma \tilde{v}^{(H)}(s'_H)$. Here, 
for each $s \in \mS$ and $a \in [K]$, the reward $\mR(s, a)$ is an independent, bounded random variable taking
value in $[-R_{\max}, R_{\max}]$ with distribution dependent on $s, a$; $\tilde{v}^{(H)}(\cdot)$ is the input of value function proxy 
to the MCTS algorithm denoted as $\hat{V}(\cdot)$, and $ \gamma \in [0,1)$ is the discount factor. Recall that 
$\varepsilon_0 = \|\hat{V} - V^*\|_\infty$ and $\|V^*\|_\infty \leq V_{\max}$. Therefore, $\|\tilde{v}^{(H)}\|_\infty = \|\hat{V}\|_\infty \leq V_{\max} + \varepsilon_0$, and the reward collected at node $s_{i, H-1}$ by following any action is bounded, in absolute value, by 
$\tilde{R}_{\max}^{(H-1)} = R_{\max} + \gamma(V_{\max} + \varepsilon_0)$.


As part of the MCTS algorithm as described in \eqref{eq:alg_mcts}, when node $s_{i, H-1}$ is visited for the $t+1$ time with $t \geq 0$, the action taken is 
\begin{align*}
  \arg\max_{a\in\mathcal{A}}& \bigg\{ \frac{1}{u_a} \sum_{j=1}^{u_a} \Big(r(s_{i, H-1}, a)(j) + \gamma \tilde{v}^{(H)}(s_{i, H-1} \circ a)(j)\Big) 
  + 
\frac{\big(\beta^{(H)}\big)^{1/\xi^{(H)}}\cdot \big(t\big)^{\alpha^{(H)}/\xi^{(H)}}}{\big(u_a\big)^{1-\eta^{(H)}}}\bigg\},    
\end{align*}
where $u_a \leq t$ is the number of times action $a$ has been chosen thus far at state $s_{i, H-1}$ in the $t$ visits so far, $r(s_{i, H-1}, a)(j)$ is
the $j$th sample of random variable per distribution $\mR(s_{i, H-1}, a)$, and $\tilde{v}^{(H)}(s_{i, H-1} \circ a)(j)$ is the reward evaluated at
leaf node $s_{i, H-1} \circ a$ for the $j$th time. Note that for all $j$, the reward evaluated at leaf node $s_{i, H-1} \circ a$ is the same and equals to 
$\tilde{v}^{(H)}(\cdot)$, the input value function proxy for the algorithm. When $u_a = 0$, we use notation $\infty$ to represent quantity inside
the $\arg\max$. The net discounted reward collected by node $s_{i, H-1}$ during its total of $t \geq 1$ visits is simply the sum of rewards obtained
by selecting the actions per the policy -- which includes the reward associated with taking an action and the evaluation of $\tilde{v}^{(H)}(\cdot)$ for appropriate
leaf node, discounted by $\gamma$. In effect, at each node $s_{i, H-1}$, we are using the UCB policy described in Section \ref{sec:nonstatmab} with parameters $\alpha^{(H)}, \beta^{(H)}, \xi^{(H)}, \eta^{(H)}$ with $K$ possible actions, where
the rewards collected by playing any of these $K$ actions each time is simply the summation of bounded independent and identical (for a given action) 
random variable and a deterministic evaluation. By applying Lemma \ref{lem.prelim}, where $X$s correspond to independent rewards, $\rho = \gamma$,
and $Y$s correspond to deterministic evaluations of $\tilde{v}^{(H)}(\cdot)$, we obtain that for given $\xi^{(H)} > 0$ and $\eta^{(H)} \in [\frac12, 1)$, 
there exists $\beta^{(H)}$ such that the collected rewards at $s_{i,H-1}$ (i.e., sum of i.i.d.\ reward and deterministic evaluations) satisfy the convergence property cf. \eqref{eq:nonstmab.conv} and 
concentration property cf. \eqref{eq:nonstmab.conc} stated in Section \ref{sec:nonstatmab}. Therefore, by an application of Theorem \ref{thm:non.stat.mab}, 
we conclude Lemma \ref{lem:leaf} stated below. We define some notations first:
\begin{align}
\mu^{(H-1)}_a(s_{i, H-1}) & = \mathbb{E}[\mR(s_{i, H-1}, a)] + \gamma \tilde{v}^{(H)}(s_{i, H-1} \circ a), \nonumber \\
\mu^{(H-1)}_*(s_{i, H-1}) & = \max_{a \in [K]} \mu^{(H-1)}_a(s_{i, H-1})\nonumber \\ 
a^{(H-1)}_*(s_{i, H-1}) & \in \arg\max _{a \in [K]} \mu^{(H-1)}_a(s_{i, H-1}) \label{eq:lemma5_def} \\
\Delta^{(H-1)}_{\min}(s_{i, H-1}) & = \mu^{(H-1)}_*(s_{i, H-1}) - \max_{a \neq a^{(H-1)}_*(s_{i, H-1})} \mu^{(H-1)}_a(s_{i,H-1}).\nonumber 
\end{align}
We shall assume 
that the maximizer in the set $\arg\max _{a \in [K]} \mu^{(H-1)}_a(s_{i, H-1})$ is unique, i.e. $\Delta_{\min}^{(H-1)}(s_{i, H-1}) > 0$. And further note that 
all rewards belong to $[-\tilde{R}_{\max}^{(H-1)}, \tilde{R}_{\max}^{(H-1)}]$. 
 
\begin{lem}\label{lem:leaf}
Consider a node corresponding to state $s_{i, H-1}$ at level $H-1$ within the MCTS for $i \in [n_{H-1}]$. Let 
$\tilde{v}^{(H-1)}(s_{i, H-1})_n$ be the total discounted reward collected at $s_{i, H-1}$ during $n \geq 1$ visits of it, to one
of its $K$ leaf nodes under the UCB policy. Then, for the choice of appropriately large $\beta^{(H)} > 0$,  for a given $\xi^{(H)} > 0$, $\eta^{(H)} \in [\frac12, 1)$
and $\alpha^{(H)} > 2$, we have
\begin{enumerate}
    \item[A.] Convergence:
    \begin{align*}
        &\left|\E\Big[\frac1n \tilde{v}^{(H-1)}(s_{i, H-1})_n\Big]- \mu^{(H-1)}_*(s_{i, H-1}) \right|\\
        &\leq \frac{2\blue{\tilde{R}_{\max}^{(H-1)}}(K-1)\cdot\Big(\big(
        \frac{2(\beta^{(H)})^{\frac{1}{\xi^{(H)}}}}{\Delta^{(H-1)}_{\min}(s_{i, H-1})} \big)^{\frac{1}{1-\eta^{(H)}}}\cdot n^{\frac{\alpha^{(H)}}{\xi^{(H)}(1-\eta^{(H)})}}+\frac{2}{\alpha^{(H)}-2} + 1\Big)}{n}.
    \end{align*}
    \item[B.] Concentration: there exist constants, $\beta'>1$ and $\xi'>0$ and $1/2\leq\eta'<1$ such that for every $n\geq 1$ and every $z\geq 1$,
\begin{align*}
&\P\big(\tilde{v}^{(H-1)}(s_{i, H-1})_n-n\mu^{(H-1)}_*(s_{i, H-1})\geq n^{\eta'} z\big) \leq\frac{\beta'}{z^{\xi'}},\\
& \P\big(\tilde{v}^{(H-1)}(s_{i, H-1})_n-n\mu^{(H-1)}_*(s_{i, H-1}) \leq -n^{\eta'} z\big) \leq\frac{\beta'}{z^{\xi'}},
\end{align*}
where $\eta'=\frac{\alpha^{(H)}}{\xi^{(H)}(1-\eta^{(H)})}$, $\xi'=\alpha^{(H)} -1$, and $\beta'$ is a large enough constant that is function of parameters 
$\alpha^{(H)}, \beta^{(H)}, \xi^{(H)},$  $\eta^{(H)}, \tilde{R}_{\max}^{(H-1)}, K, \Delta^{(H-1)}_{\min}(s_{i, H-1})$. 
\end{enumerate}
\end{lem}
Let $\Delta_{\min}^{(H-1)} = \min_{i \in [n_{H-1}]} \Delta^{(H-1)}_{\min}(s_{i, H-1})$. Then, the rate of convergence for each node $s_{i,H-1}$, $i\in[n_{H-1}]$ can be uniformly simplified as
\begin{align*}
\delta^{(H-1)}_n=&  \frac{2\tilde{R}_{\max}^{(H-1)}(K-1)\cdot\Big(\big(
        \frac{2(\beta^{(H)})^{\frac{1}{\xi^{(H)}}}}{\Delta_{\min}^{(H-1)}} \big)^{\frac{1}{1-\eta^{(H)}}}\cdot n^{\frac{\alpha^{(H)}}{\xi^{(H)}(1-\eta^{(H)})}}+\frac{2}{\alpha^{(H)}-2} + 1\Big)}{n} \nonumber \\
         =&~\Theta\Big(n^{\frac{\alpha^{(H)}}{\xi^{(H)}(1-\eta^{(H)})} - 1}\Big)\nonumber \\
        \stackrel{(a)}{=}&~O\big(n^{\eta - 1}\big), \nonumber
\end{align*}
where (a) holds since $\alpha^{(H)} = \xi^{(H)}(1-\eta^{(H)})\eta^{(H)}, ~\eta^{(H)} = \eta$.
It is worth remarking that $\mu^{(H-1)}_*(s_{i, H-1})$, as defined in (\ref{eq:lemma5_def}),  is precisely the value function estimation for $s_{i, H-1}$ 
at the end of one step of value iteration starting with $\hat{V}$.


\subsection{Recursion: Going From Level $h$ to $h-1$.}\label{ssec:part1.2} 

Lemma \ref{lem:leaf} suggests that the necessary assumption of Theorem \ref{thm:non.stat.mab}, i.e. \eqref{eq:nonstmab.conv} and \eqref{eq:nonstmab.conc}, 
are satisfied by $\tilde{v}^{(H-1)}_n$ for each node or state at level $H-1$, with $\alpha^{(H-1)}, \xi^{(H-1)}, \eta^{(H-1)}$ as defined per relationship \eqref{eq:mcts_sequence_1} - \eqref{eq:mcts_sequence_4} and with appropriately defined large enough constant $\beta^{(H-1)}$. We shall argue that result
similar to Lemma \ref{lem:leaf}, but for node at level $H-2$, continues to hold with parameters $\alpha^{(H-2)}, \xi^{(H-2)}, \eta^{(H-2)}$ as defined per relationship \eqref{eq:mcts_sequence_1} - \eqref{eq:mcts_sequence_4} and with appropriately defined large enough constant $\beta^{(H-2)}$. 
And similar argument will continue to apply going from level $h$ to $h-1$ for all $h \leq H-1$. That is, we shall assume that the necessary assumption of Theorem \ref{thm:non.stat.mab}, i.e. \eqref{eq:nonstmab.conv} 
and \eqref{eq:nonstmab.conc}, holds for $\tilde{v}^{(h)}(\cdot)$, for all nodes at level $h$ with $\alpha^{(h)}, \xi^{(h)}, \eta^{(h)}$ as defined per relationship \eqref{eq:mcts_sequence_1} - \eqref{eq:mcts_sequence_4} and with appropriately defined large enough constant $\beta^{(h)}$, and then argue that such holds for 
nodes at level $h-1$ as well. This will, using mathematical induction, allow us to prove the results for all $h\geq 1$.

To that end, consider any node at level $h-1$. Let there be $n_{h-1}$ nodes at level $h-1$ corresponding to states 
$s_{1, h-1},\dots, s_{n_{h-1}, h-1} \in \mS$. 
Consider a node corresponding to state $s_{i, h-1}$ at level $h-1$ within the MCTS for $i \in [n_{h-1}].$
%
%
%
As part of the algorithm, whenever this node is visited, one of the $K$ feasible action is taken. 
When an action $a \in [K]$ is taken, the node $s'_{h} = s_{i, h-1} \circ a$, at the  level $h$ is reached. This results 
in reward at node $s_{i, h-1}$ at level $h-1$ being equal to $\mR(s_{i, h-1}, a) + \gamma \tilde{v}^{(h)}(s'_h)$. As noted before, 
$\mR(s, a)$ is an independent, bounded valued random variable while $\tilde{v}^{(h)}(\cdot)$ is effectively collected by following
a path all the way to the leaf level. Inductively, we assume that $\tilde{v}^{(h)}(\cdot)$ satisfies the convergence and concentration property for each node or state at level 
$h$, with $\alpha^{(h)}, \xi^{(h)}, \eta^{(h)}$ as defined per relationship \eqref{eq:mcts_sequence_1} - \eqref{eq:mcts_sequence_4} and
with appropriately defined large enough constant $\beta^{(h)}$.  Therefore, by an application of Lemma \ref{lem.prelim}, 
it follows that this combined reward continues to  satisfy \eqref{eq:nonstmab.conv} and \eqref{eq:nonstmab.conc}, with 
$\alpha^{(h)}, \xi^{(h)}, \eta^{(h)}$ as defined per relationship \eqref{eq:mcts_sequence_1} - \eqref{eq:mcts_sequence_4} and with 
a  large enough constant which we shall denote as $\beta^{(h)}$. These constants are used by the MCTS policy.  
By an application of Theorem~\ref{thm:non.stat.mab}, we can obtain the following Lemma~\ref{lem:recurse} {regarding the convergence and concentration properties for the reward sequence collected at node $s_{i,h-1}$ at level $h-1.$ Similar to the notation in Eq.~(\ref{eq:lemma5_def}), let
\begin{align}
\mu^{(h-1)}_a(s_{i, h-1}) & = \mathbb{E}[\mR(s_{i, h-1}, a)] + \gamma \mu^{(h)}_*(s_{i, h-1} \circ a) \nonumber \\
\mu^{(h-1)}_*(s_{i, h-1}) & = \max_{a \in [K]} \mu^{(h-1)}_a(s_{i, h-1}) \nonumber \\
a^{(h-1)}_*(s_{i, h-1}) & \in \arg\max _{a \in [K]} \mu^{(h-1)}_a(s_{i, h-1})  \label{eq:induct.not2} \\ 
\Delta^{(h-1)}_{\min}(s_{i, h-1}) & = \mu^{(h-1)}_*(s_{i, h-1}) - \max_{a \neq a^{(h-1)}_*(s_{i, h-1})} \mu^{(h-1)}_a(s_{i, h-1}).\nonumber
\end{align}}
Again, we shall assume that the maximizer in the set $\arg\max _{a \in [K]} \mu^{(h-1)}_a(s_{i, h-1})$ is unique, i.e. $\Delta^{(h-1)}_{\min}(s_{i, h-1}) > 0$. 
Define
 $\tilde{R}_{\max}^{(h-1)} = R_{\max} + \gamma \tilde{R}_{\max}^{(h)},$ where $\tilde{R}^{(H)}=V_{\max}+\varepsilon_0.$ Note that all rewards collected at level $h-1$ belong to $[-\tilde{R}^{(h-1)}_{\max}, \tilde{R}^{(h-1)}_{\max}]$. 

\begin{lem}\label{lem:recurse}
Consider a node corresponding to state $s_{i, h-1}$ at level $h-1$ within the MCTS for $i \in [n_{h-1}]$. Let 
$\tilde{v}^{(h-1)}(s_{i, h-1})_n$ be the total discounted reward collected at $s_{i, h-1}$ during $n \geq 1$ visits. Then, 
for the choice of appropriately large $\beta^{(h)} > 0$, for a given $\xi^{(h)} > 0$, $\eta^{(h)} \in [\frac12, 1)$
and $\alpha^{(h)} > 2$, we have
\begin{enumerate}
    \item[A.] Convergence:
    \begin{align*}
        &\left|\E\Big[\frac1n \tilde{v}^{(h-1)}(s_{i, h-1})_n\Big]- \mu^{(h-1)}_*(s_{i, h-1}) \right|\\
        &\leq \frac{2\tilde{R}_{\max}^{(h-1)}(K-1)\cdot\Big(\big(
        \frac{2(\beta^{(h)})^{\frac{1}{\xi^{(h)}}}}{\Delta^{(h-1)}_{\min}(s_{i, h-1})} \big)^{\frac{1}{1-\eta^{(h)}}}\cdot n^{\frac{\alpha^{(h)}}{\xi^{(h)}(1-\eta^{(h)})}}+\frac{2}{\alpha^{(h)}-2} + 1\Big)}{n}.
    \end{align*}
    \item[B.] Concentration: there exist constants, $\beta'>1$ and $\xi'>0$ and $1/2\leq\eta'<1$ such that for  $n\geq 1$, 
   $z\geq 1$,
\begin{align*}
&\P\big(\tilde{v}^{(h-1)}(s_{i, h-1})_n-n\mu^{(h-1)}_*(s_{i, h-1})\geq n^{\eta'} z\big) \leq\frac{\beta'}{z^{\xi'}},\\
& \P\big(\tilde{v}^{(h-1)}(s_{i, h-1})_n-n\mu^{(h-1)}_*(s_{i, h-1}) \leq -n^{\eta'} z\big) \leq\frac{\beta'}{z^{\xi'}},
\end{align*}
where $\eta'=\frac{\alpha^{(h)}}{\xi^{(h)}(1-\eta^{(h)})}$, $\xi'=\alpha^{(h)} -1$, and $\beta'$ is a large enough constant that is function of parameters 
$\alpha^{(h)}, \beta^{(h)}, \xi^{(h)},$  $\eta^{(h)}, \tilde{R}_{\max}^{(h-1)}, K, \Delta^{(h-1)}_{\min}(s_{i, h-1})$. 
\end{enumerate}
\end{lem}
As before, let us define $\Delta_{\min}^{(h-1)}  = \min_{i \in [n_{h-1}]} \Delta^{(h-1)}_{\min}(s_{i, h-1})$. 
Similarly, we can show that for every node $s_{i,h-1}$, $i\in[n_{h-1}]$, the rate of convergence in Lemma \ref{lem:recurse} can be uniformly simplified as
\begin{align*}\label{eq:recurse.1}
\delta^{(h-1)}_n 
& =  \frac{2\tilde{R}_{\max}^{(h-1)}(K-1)\cdot\Big(\big(
        \frac{2(\beta^{(h)})^{\frac{1}{\xi^{(h)}}}}{\Delta_{\min}^{(h-1)}} \big)^{\frac{1}{1-\eta^{(h)}}}\cdot n^{\frac{\alpha^{(h)}}{\xi^{(h)}(1-\eta^{(h)})}}+\frac{2}{\alpha^{(h)}-2} + 1\Big)}{n} \nonumber \\
        & ~=~\Theta\Big(n^{\frac{\alpha^{(h)}}{\xi^{(h)}(1-\eta^{(h)})} - 1}\Big) 
        ~=~O\big(n^{\eta - 1}\big),
\end{align*}
where the last equality holds as $~\alpha^{(h)} = \xi^{(h)}(1-\eta^{(h)})\eta^{(h)}$ and $~\eta^{(h)} = \eta.$  Again, it is worth remarking, inductively, that $\mu^{(h-1)}_*(s_{i, h-1})$ is precisely the value function estimation for 
$s_{i, h-1}$ at the end of $H-h+1$ steps of value iteration starting with $\hat{V}$.

\smallskip
\noindent {\em {\bf Remark} (Recursive Relation among Parameters)}. 
{With the above development, we are ready to elaborate our choice of parameters in Theorem \ref{thm:MCTS}, defined recursively via Eqs.~(\ref{eq:mcts_sequence_1})-(\ref{eq:mcts_sequence_4}). In essence, those parameter requirements originate from our analysis of the non-stationary MAB, i.e., Theorem \ref{thm:non.stat.mab}. Recall that from our previous analysis, the key to establish the 
MCTS guarantee is to recursively argue the convergence and the polynomial concentration properties at each level; that is, we recursively solve the non-stationary MAB problem at each level. In order to do so, we apply our result on the non-stationary MAB (Theorem \ref{thm:non.stat.mab}) recursively at each level. Importantly, recall that Theorem \ref{thm:non.stat.mab} only holds when $\xi\eta(1-\eta)\leq \alpha< \xi(1-\eta)$ and $\alpha > 2$, under which it leads to the recursive conclusions $\eta' = \frac{\alpha}{\xi(1-\eta)}$ and $\xi' = \alpha - 1$. Using our notation with superscript indicating the levels, this means that apart from the parameters at the leaf level (level $H$) which could be freely chosen, we must choose parameters of other levels recursively so that the following conditions hold:
\begin{align*}
     &\alpha^{(h)}>2, \quad  \xi^{(h)}\eta^{(h)}(1-\eta^{(h)})\leq\alpha^{(h)}<\xi^{(h)}(1-\eta^{(h)}), \\
    &\xi^{(h)} =\alpha^{(h+1)}-1\textrm{ and } \eta^{(h)}=\frac{\alpha^{(h+1)}}{\xi^{(h+1)}(1-\eta^{(h+1)})}.
\end{align*}
It is not hard to see that the conditions in Theorem \ref{thm:MCTS} guarantee the above. There might be other sequences of parameters satisfying the requirements, but our particular choice gives cleaner  
analysis as presented in this paper.
}

%
%


\subsection{Error Analysis for Value Function Iteration}\label{ssec:part2}
We now move to the second part of the proof. 
The value function iteration improves the estimation of optimal value function by iterating Bellman equation. In effect, the MCTS tree 
is ``unrolling" $H$ steps of such an iteration. Precisely, let $V^{(h)}(\cdot)$ denote the value function after $h$ iterations starting
with $V^{(0)} = \hat{V}$. By definition, for any $h \geq 0$ and $s \in \mS$, 
\begin{align}
V^{(h+1)}(s) & = \max_{a \in [K]} \Big(\mathbb{E}[\mR(s, a)] + \gamma V^{(h)}(s \circ a)\Big). 
\end{align}
Recall that value iteration is contractive with respect to $\|\cdot\|_\infty$ norm (cf. \cite{bertsekas2017dynamic}). That is, for any $h \geq 0$, 
\begin{align}\label{eq:value.iter.1}
\| V^{(h+1)} - V^*\|_\infty & \leq \gamma \| V^{(h)} - V^*\|_\infty.
\end{align}
As remarked earlier, $\mu^{(h-1)}_*(s_{i, h-1})$, the mean reward collected at node $s_{i, h-1}$ for $i \in [n_{h-1}]$ for any $h \geq 1$, is 
precisely $V^{(H-h+1)}(s_{i, h-1})$ starting with $V^{(0)} = \hat{V}$, the input to MCTS policy. Therefore, the mean reward collected at root
node $s^{(0)}$ of the MCTS tree satisfies $\mu^{(0)}_*(s^{(0)}) = V^{(H)}(s^{(0)})$. Using \eqref{eq:value.iter.1}, we obtain the following Lemma. 
\begin{lem}\label{lem:value.iter}
The mean reward collected under the MCTS policy at root note $s^{(0)}$,  $\mu^{(0)}_*(s^{(0)})$, starting with input value function proxy 
$\hat{V}$ is such that 
\begin{align}
|\mu^{(0)}_*(s^{(0)}) - V^*(s^{(0)})| & \leq \gamma^H \|\hat{V} - V^*\|_\infty. 
\end{align}
\end{lem}

\subsection{Completing Proof of Theorem~\ref{thm:MCTS}}\label{ssec:conc}

In summary, using Lemma \ref{lem:recurse}, we conclude that the recursive relationship going from level $h$ to $h-1$ 
holds for all $h \geq 1$ with level $0$ being the root. At root $s^{(0)}$, the query  state that is input to the MCTS policy, 
we have that after $n$ total simulations of MCTS, the empirical average of the rewards over these 
$n$ trial, $\frac{1}{n} \tilde{v}^{(0)}(s_0)_n$ is such that (using the fact that $\alpha^{(0)} = \xi^{(0)}(1-\eta^{(0)}) \eta^{(0)}$)
\begin{align}\label{eq.final.1}
\left| \mathbb{E}\Big[\frac{1}{n} \tilde{v}^{(0)}(s_0)_n\Big] - \mu^{(0)}_* \right| & = O\Big(n^{\frac{\alpha^{(0)}}{\xi^{(0)}(1-\eta^{(0)})} - 1}\Big)~=~O\Big(n^{\eta - 1}\Big), 
\end{align}
where $\mu^{(0)}_*$ is the value function estimation for $s^{(0)}$ after $H$ iterations of value function iteration starting with $\hat{V}$. By Lemma \ref{lem:value.iter}, 
we have 
\begin{align}\label{eq.final.2}
|\mu^{(0)}_* - V^*(s^{(0)})| & \leq \gamma^H \varepsilon_0,
\end{align}
since $\varepsilon_0 = \|\hat{V} - V^*\|_\infty$. Combining \eqref{eq.final.1} and \eqref{eq.final.2}, 
\begin{align}\label{eq.final.3}
\left| \mathbb{E}\Big[\frac{1}{n} \tilde{v}^{(0)}(s_0)_n\Big] - V^*(s^{(0)}) \right| & \leq \gamma^H \varepsilon_0 + O\Big(n^{\eta - 1}\Big).
\end{align}
This concludes the proof of Theorem \ref{thm:MCTS}.

\subsection{Proof of Lemma~\ref{lem.prelim}} \label{appx:proof_prelim}

The convergence property, $\lim_{n\to\infty} \E[\bar{Z}_{n}]  = \mu_X + \rho \mu_Y$, follows simply by linearity of expectation. 
For concentration, consider the following: since $X$s are i.i.d. bounded random variables taking value in $[-B, B]$, 
by Hoeffding's inequality \citep{Hoeffding}, we have that for $t \geq 0$, 
\begin{align}\label{eq:h1}
&\P\big(n \bar{X}_n - n \mu_X \geq n t \big) \leq \exp\Big(-\frac{t^2 n}{2B^2}\Big), \\
&\P\big(n \bar{X}_n - n \mu_X \leq - n t \big)  \leq \exp\Big(-\frac{t^2 n}{2B^2}\Big). \nonumber
\end{align}
Therefore, 
\begin{align}
\P\Big(n\bar{Z}_{n}-n(\mu_{X} + \rho \mu_{Y}) \geq n^{\eta} z\big)
& \leq \P\Big(n\bar{X}_{n}-n\mu_{X} \geq \frac{n^{\eta}z}{2} \Big) + 
\P\Big(n\bar{Y}_{n}-n\mu_{Y} \geq \frac{n^{\eta}z}{2\rho} \Big) \nonumber \\
& \leq \exp\Big(-\frac{z^2 n^{2\eta -1} }{8B^2}\Big) + \frac{\beta 2^\xi \rho^\xi }{z^{\xi}}  \nonumber \\
& \leq \frac{\beta'}{z^\xi}, 
\end{align}
where $\beta'$ is a large enough constant depending upon $\rho, \xi, \beta$ and $B$. The other-side of the inequality follows similarly.
This completes the proof.

%% file: proof_sl.tex
\section{Proof of Theorem \ref{thm:MCTS_SL_deterministic}} \label{sec:proof.rl}

First, we establish a useful property of nearest neighbor supervised learning presented in Section \ref{ssec:sl.policy}. This is stated in Section \ref{ssec:proof.sl}. 
We will use it, along with the guarantees obtained for MCTS in Theorem \ref{thm:MCTS} to establish Theorem \ref{thm:MCTS_SL_deterministic} in Section \ref{ssec:proof.rl}. 
Throughout, we shall assume the setup of Theorem \ref{thm:MCTS_SL_deterministic}. 

\subsection{Guarantees for Supervised Learning}\label{ssec:proof.sl}

Let $\delta \in (0,1)$ be given. As stated in Section \ref{ssec:sl.policy},  let 
$K(\delta, d) = \Theta(\delta^{-d})$ be the collection of balls of radius $\delta$, say 
$c_i, ~i \in [K(\delta, d)]$, so that they cover $\mS$, i.e. $\mS \subset \cup_{i \in [K(\varepsilon, d)]} c_i$. 
Also, by construction, each of these balls have intersection with $\mS$ whose volume is at least $C_d \delta^d$. 
Let $S = \{s_i: i \in [N] \}$ denote $N$ state samples from $\mS$ uniformly at random and independent of each other. 
For each state $s \in \mS$, let $V: \mS \to [-{V_{\max}, V_{\max}}]$ be such that $| \mathbb{E}[V(s)] - V^*(s)| \leq \Delta$. 
Let the nearest neighbor supervised learning described in Section \ref{ssec:sl.policy} produce estimate 
$\hat{V}: \mS \to \mathbb{R}$ using labeled data points $(s_i, V(s_i))_{i \in [N]}$. Then, we claim the following
guarantee. Proof can be found in Section \ref{appx:proof_sl_final}. 
\begin{lem}\label{lem:sl.final}
Under the above described setup, as long as $N \geq 32 \max(1, \delta^{-2} V_{\max}^2) C_d^{-1} \delta^{-d}\log \frac{K(\delta,d)}{\delta}$, 
i.e., $N = \Omega(d \delta^{-d-2} \log \delta^{\blue{-1}})$, 
\begin{align}
\mathbb{E}\big[\sup_{s \in \mS} | \hat{V}(s) - V^*(s)|\big] & \leq \Delta + (C+1) \delta + \frac{{4 V_{\max}}\delta^2}{K(\delta, d)}.
\end{align}
 \end{lem}

\subsection{Establishing Theorem \ref{thm:MCTS_SL_deterministic}}\label{ssec:proof.rl}

Using Theorem \ref{thm:MCTS} and Lemma \ref{lem:sl.final}, we complete the proof of Theorem \ref{thm:MCTS_SL_deterministic} under appropriate choice of algorithmic parameters. We start by setting some notation. 

To that end, the algorithm as described in Section \ref{ssec:rl.policy} iterates between MCTS and supervised learning. In particular, let $\ell\geq 1$ denote the iteration index. Let $m_\ell$ be the number of states that are sampled uniformly at random, independently,  over $\mS$ in this iteration, denoted as $S^{(\ell)} = \{s_i^{(\ell)}: i \in [m_\ell]\}$. Let $V^{(\ell-1)}$ be the input of value function from prior iteration, using which the MCTS algorithm with $n_\ell$ simulations obtains improved estimates of value function for states in $S^{(\ell)}$ denoted as $\hat{V}^{(\ell)}(s_i^{(\ell)}), ~i \in [m_\ell]$. Using $(s_i^{(\ell)}, \hat{V}^{(\ell)}(s_i^{(\ell)}))_{i \in [m_\ell]}$, the nearest neighbor supervised learning as described above with balls of appropriate radius 
$\delta_\ell \in (0,1)$ produces estimate $V^{(\ell)}$ for all states in $\mS$. Let $\mF^{(\ell)}$ denote the smallest 
$\sigma$-algebra containing all information pertaining to the algorithm (both MCTS and supervised learning). Define the error
under MCTS in iteration $\ell$ as 
\begin{align}\label{eq:mcts.err}
\emcts^{(\ell)} & = \E\big[\sup_{s\in \mS}\big|\E\big[\hat{V}^{(\ell)}(s) \big| \mF^{(\ell-1)} \big]-V^{*}(s)\big|\big].
\end{align}
And, the error for supervised learning in iteration $\ell$ as 
\begin{align}
\tsl^{(\ell)} & = \sup_{s \in \mS} \big| V^{(\ell)}(s) - V^*(s) \big|, ~\text{and}~\esl^{(\ell)} = \mathbb{E}\big[\tsl^{(\ell)}\big]. 
\end{align}
Recall that in the beginning, we set $V^{(0)}(s) = 0$ for all $s \in \mS$. Since $V^*(\cdot) \in [-V_{\max}, V_{\max}]$, we have
that $\esl^{(0)} \leq V_{\max}$. {Further, it is easy to see that if the leaf estimates (i.e., the output of the supervised learning from the previous iteration) is bounded in $[-V_{\max}, V_{\max}]$, then the output of the MCTS algorithm is always bounded in $[-V_{\max}, V_{\max}]$. That is, since $V^{(0)}(s) = 0$ and the nearest neighbor supervised learning produces estimate $V^{(l)}$ via simple averaging, inductively, the output of the MCTS algorithm is always bounded in $[-V_{\max}, V_{\max}]$ throughout every iteration.}

With the notation as set up above, it follows that for a given $\delta_\ell \in (0,1)$ with $m_\ell$ satisfying condition of Lemma \ref{lem:sl.final}, i.e. $m_\ell = \Omega(d \delta_\ell^{-d-2} \log \delta_\ell^{\blue{-1}})$, and with the nearest neighbor supervised learning using
$\delta_\ell$ radius balls for estimation, we have the following recursion: 
\begin{align}\label{eq:recurse.1p}
\esl^{(\ell)} & \leq \emcts^{(\ell)} + (C+1) \delta_\ell + \frac{{4 V_{\max}}\delta_\ell^2}{K(\delta_\ell, d)} 
~ \leq ~\emcts^{(\ell)} + C' \delta_\ell, 
\end{align}
where $C'$ is a large enough constant, since $\frac{\delta_\ell^2}{K(\delta_\ell, d)} = 
\Theta(d \delta_\ell^{d + 2})$ which is $O(\delta_\ell)$ for all $\delta_\ell \in (0,1)$. By Theorem \ref{thm:MCTS}, for 
iteration $\ell+1$ that uses the output of supervised learning estimate, $V^{(\ell)}$, as the input to the MCTS algorithm, we obtain 
\begin{align}
\big|\E\big[\hat{V}^{(\ell+1)}(s)\big|\mF^{(\ell)}\big]-V^{*}(s)\big|\leq\gamma^{H^{(\ell+1)}}\E\big[\tsl^{(\ell)}\big | \mF^{(\ell)}\big] + O\big(n_{\ell +1}^{\eta-1}\big), \forall s\in \mS, \label{eq:single_mcts_error_sl}
\end{align}
where $\eta\in[1/2,1)$ is the constant utilized by MCTS with fixed height of tree being $H^{(\ell+1)}$. 
This then implies that 
\begin{align}
\emcts^{(\ell+1)} & = 
\E\big[\sup_{s\in \mS}\big|\E\big[\hat{V}^{(\ell+1)}(s)\big|\mF^{(\ell)}\big]-V^{*}(s)\big|\big] \nonumber \\
 &\leq  \gamma^{H^{(\ell+1)}}\E\Big[\E\big[\tsl^{(\ell)} \big| \mF^{(\ell)}\big]\Big] + O\big(n_{\ell+1}^{\eta-1}\big) \nonumber \\
& \leq \gamma^{H^{(\ell+1)}} \Big(\emcts^{(\ell)} + C' \delta_\ell\Big) + O\big(n_{\ell+1}^{\eta-1}\big). \label{eq:single_mcts_error_mcts}
\end{align}

{Denote by $\lambda\triangleq (\frac{\varepsilon}{V_{\max}})^{1/L}$. Note that since the final desired error $\varepsilon$ should be less than $V_{\max}$ (otherwise, the problem is trivial by just outputing $0$ as the final estimates for all the states), we have $\lambda < 1$. Let us set the algorithmic parameters for MCTS and nearest neighbor supervised learning as follows: for each $\ell\geq 1$,
\begin{align}
H^{(\ell)} =\big\lceil \log_{\gamma}\frac{\lambda}{8}\big\rceil,
\delta_{\ell} =\frac{3V_{\max}}{4C'}\lambda^{\ell},  
n_\ell=\kappa_l \Big(\frac{8}{V_{\max}\lambda^{\ell}}\Big)^{\frac{1}{1-\eta}}, \label{eq:mcts_sl_parameters}
\end{align}
where $\kappa_l>0$ is a sufficiently large constant such that $O\big(n_{\ell}^{\eta-1}\big)=\frac{V_{\max}}{8}\lambda^{\ell}$. 
Substituting these values into Eq. (\ref{eq:single_mcts_error_mcts}) yields
\[
\varepsilon_{\text{mcts}}^{(\ell+1)}=\E\Big[\sup_{s\in \mS}\big|\E\big[\hat{V}^{(\ell+1)}(s)|\mathcal{F}^{(\ell)}\big]-V^{*}(s)\big|\Big]\leq\frac{\lambda}{8}\varepsilon_{\text{mcts}}^{(\ell)}+\frac{7V_{\max}}{32}\lambda^{\ell+1}.
\]
Note that by (\ref{eq:single_mcts_error_sl}) and (\ref{eq:mcts_sl_parameters}), and the fact that $\varepsilon_{\text{sl}}^{(0)}\leq V_{\max}$, we have
\[
\varepsilon_{\text{mcts}}^{(1)}\leq \frac{\lambda}{8}\varepsilon_{\text{sl}}^{(0)} + \frac{\lambda}{8}V_{\max}\leq\frac{\lambda}{4}V_{\max}.
\]
It then follows inductively that
\begin{align*}
\varepsilon_{\text{mcts}}^{(\ell)} & \leq\lambda^{\ell-1}\varepsilon_{\text{mcts}}^{(1)}=\frac{V_{\max}}{4}\lambda^{\ell}.
\end{align*}
As for the supervised learning oracle, $\forall s\in\mS,$ Eq. \eqref{eq:recurse.1p} implies
\begin{align*}
\E\big[\sup_{s\in\mathcal{S}}\big\vert V^{(\ell)}(s) -V^{*}(s)\big\vert\big]
\leq \varepsilon_{\text{mcts}}^{(\ell)}+\frac{3V_{\max}}{4}\lambda^{\ell}\leq V_{\max}\lambda^{\ell}.
\end{align*}
This implies that
\begin{align*}
\E\big[\sup_{s\in\mathcal{S}}\big\vert V^{(L)}(s) -V^{*}(s)\big\vert\big]
\leq V_{\max}\lambda^{L} = \varepsilon.
\end{align*}
We now calculate the sample complexity, i.e., the total number of state transitions required for the algorithm. During the $\ell$-th iteration, each query of MCTS oracle requires $n_\ell$ simulations. Recall that the number of querying MCTS oracle, i.e., the size of training
set $\mS^{(\ell)}$ for the nearest neighbor supervised step, should satisfy $m_\ell=\Omega(d\delta_\ell^{-d-2}\log\delta_\ell^{-1})$ (cf. Lemma \ref{lem:sl.final}).
From Eq. (\ref{eq:mcts_sl_parameters}), we have
\[
H^{(\ell)}= c'_0 \log{\lambda^{-1}}, \quad \delta^{(\ell)}=c_1\lambda^\ell, \text{and}\quad n_\ell=c'_2\lambda^{-\ell/(1-\eta)},
\]
where $c'_0,c_1,c'_2,$ are constants independent of $\lambda$ and $\ell.$
Note that each simulation of MCTS samples $H^{(\ell)}$ state transitions.  
Hence, the number of state transitions at the $\ell$-th iteration is given by 
\begin{align*}
M^{(\ell)} & =m_\ell n_\ell H^{(\ell)}.
\end{align*}
Therefore, the total number of state transitions after $L$ iterations is
\begin{align*}
\sum_{l=1}^{L}M^{(\ell)} & =\sum_{\ell=1}^{L}m_{\ell}\cdot n_{\ell}\cdot H^{(\ell)} = O\Big(\varepsilon^{-\big(2 + 1/(1-\eta) +d\big)}\cdot\big(\log\frac{1}{\varepsilon}\big)^5\Big).
\end{align*}
That is, for optimal choice of $\eta = 1/2$, the total number of state transitions is $O\big(\varepsilon^{-(4 +d)}\cdot\big(\log\frac{1}{\varepsilon}\big)^5\big).$
\bigskip

\subsection{Proof of Lemma~\ref{lem:sl.final}} \label{appx:proof_sl_final}

Given $N$ samples $s_i, i \in [N]$ that are sampled independently and uniformly at random over $\mS$, and 
given the fact that each ball $c_i, ~ i\in [K(\delta, d)]$ has at least $C_d \delta^d$ volume shared with $\mS$, each
of the sample falls within a given ball with probability at least $C_d \delta^d$. Let $N_i, ~ i\in [K(\delta, d)]$ denote 
the number of samples amongst $N$ samples in ball $c_i$. 

Now the number of samples falling in any given ball is lower bounded by a Binomial random variable 
with parameter $N, C_d \delta^d$. By  Chernoff bound for Binomial variable with parameter $n, p$, we 
have that 
\begin{align*}
\P(B(n,p) \leq np/2) & \leq \exp\big(-\frac{np}{8}\big). 
\end{align*}
Therefore, with an application of union bound, each ball has at least $0.5 C_d \delta^d N$ 
samples with probability at least $1 - K(\delta, d) \exp\big(- C_d \delta^d N/8\big)$. That is, for 
$N = 32 \max(1, \delta^{-2} V_{\max}^2) C_d^{-1} \delta^{-d} [\log (K(\delta, d) + \log \delta^{-1}]$,  
each ball has at  least $\Gamma = 16  \max(1, \delta^{-2} {V^2_{\max}}) (\log K(\delta, d) + \log \delta^{-1})$ 
samples with probability at least $ 1 - \frac{\delta^2}{K(\delta,d) }$. Define event 
\begin{align*}
\mathcal{E}_1 & = \{ N_i \geq 16  \max(1, \delta^{-2} {V^2_{\max}}) (\log K(\delta, d) + \log \delta^{-1}), ~\forall ~i\in [K(\delta, d)]\}.
\end{align*}
Then 
\begin{align*} 
\P({\mathcal{E}}^c_1) \leq \frac{\delta^2}{K(\delta,d) }.
\end{align*}
Now, for any $s \in \mS$, the nearest neighbor supervised learning described in Section \ref{ssec:sl.policy} produces 
estimate $\hat{V}(s)$ equal to the average value of observations for samples falling in ball $c_{j(s)}$. Let $N_{j(s)}$ denote the number of samples in ball $c_{j(s)}.$ To that end, 
\begin{align*}
\left|\hat{V}(s) - V^*(s)\right| & = \left|\frac{1}{N_{j(s)}} \Big(\sum_{i: s_i \in c_{j(s)}} V(s_i) - V^*(s)\Big)\right| \nonumber \\
& = \left|\frac{1}{N_{j(s)}} \Big(\sum_{i: s_i \in c_{j(s)}} V(s_i) - \mathbb{E}[V(s_i)] \Big)\right| 
+  \left|\frac{1}{N_{j(s)}} \Big(\sum_{i: s_i \in c_{j(s)}} \mathbb{E}[V(s_i)] - V^*(s_i) \Big)\right | \nonumber \\ 
& \qquad +\left|\frac{1}{N_{j(s)}} \Big(\sum_{i: s_i \in c_{j(s)}} V^*(s_i) - V^*(s) \Big) \right|.
\end{align*}
For the first term, since for each $s_i \in c_{j(s)}$, $V(s_i)$ is produced using independent randomness via MCTS, and
since the output $V(s_i)$ is a bounded random variable, using Hoeffding's inequality, it follows that 
\begin{align*}
\P\Big(\Big|\frac{1}{N_{j(s)}} \Big(\sum_{i: s_i \in c_{j(s)}} V(s_i) - \mathbb{E}[V(s_i)] \Big)\Big| \geq \Delta_1\Big) & \leq 
2 \exp\Big(-\frac{N_{j(s)} \Delta_1^2}{8V_{\max}^2}\Big).
\end{align*}
The second term is no more than $\Delta$ 
due to the guarantee given by MCTS 
as assumed in the setup. And finally,
the third term is no more than $C \delta$ due to Lipschitzness of $V^*$. To summarize, with probability at least
$1 - 2 \exp\Big(-\frac{N_{j(s)} \Delta_1^2}{8V_{\max}^2}\Big)$, we have that 
\begin{align*}
\left|\hat{V}(s) - V^*(s)\right| & \leq \Delta_1 + \Delta + C \delta.
\end{align*}
As can be noticed, the algorithm produces the same estimate for all $s \in \mS$ such that they map to the same
ball. And there are $K(\delta, d)$ such balls. Therefore, using union bound, it follows that with probability at least
$1 - 2 K(\delta, d) \exp\Big(-\frac{(\min_{i \in [K(\delta, d)]} N_{i}) \Delta_1^2}{8V_{\max}^2}\Big)$, 
\begin{align*}
\sup_{s \in \mS} \left|\hat{V}(s) - V^*(s)\right| & \leq \Delta_1 + \Delta + C \delta.
\end{align*}
Under event $\mathcal{E}_1$, $\min_{i \in [K(\delta, d)]} N_{i} \geq 16  \max(1, \delta^{-2} {V^2_{\max}}) (\log K(\delta, d) + \log \delta^{-1})$. Therefore, under event ${\mathcal{E}}_1$, by choosing $\Delta_1 = \delta$, we have 
\begin{align*}
\sup_{s \in \mS} \left|\hat{V}(s) - V^*(s)\right| & \leq \Delta + (C+1) \delta,
\end{align*}
with probability at least $1- \frac{2\delta^2}{K(\delta, d)}$. When event ${\mathcal{E}}_1$ does not hold or the above does not hold, 
we have trivial error bound of $2 {V_{\max}}$ on the error. Therefore, we conclude that 
\begin{align*}
\mathbb{E}\Big[\sup_{s \in \mS} \left|\hat{V}(s) - V^*(s)\right|\Big] & \leq \Delta + (C+1) \delta + \frac{{4 V_{\max}}\delta^2}{K(\delta, d)}.
\end{align*}

%% file: Appendix_lower_bound.tex
\section{Proof of Proposition~\ref{prop:lower_bound}} \label{appendix:proof_lower_bound}

The recent work~\cite{shah2018qlnn} establishes a lower bound on the sample complexity for reinforcement learning algorithms on MDPs. We follow a similar argument to establish a lower bound on the sample complexity for MDPs with deterministic transitions. We provide the proof for completeness. The key idea is to connect the problem of estimating the value function to the problem of non-parametric regression, and then leveraging known minimax lower bound for the latter. In particular, we show that a class of non-parametric regression problem can be embedded in an MDP with deterministic transitions, so any algorithm for the latter can be used to solve the former. Prior work on non-parametric regression~\citep{tsybakov2009nonparm,stone1982optimal} establishes that a certain number of observations is \emph{necessary} to achieve a given accuracy using \emph{any} algorithms, hence leading to a corresponding necessary condition for the sample size of estimating the value function in an MDP problem. We now provide the details. 

\medskip
\noindent{\bf Step 1. Non-parametric regression.} Consider the following non-parametric regression problem:
Let $ \mS:=[0,1]^{d} $ and assume that we have $T$ data pairs 
$(x_{1},y_{1}),\ldots,(x_{T},y_{T})$ such that conditioned on $x_1,\ldots,x_n,$ the random variables $y_1,\ldots,y_n$ are independent and satisfy
\begin{equation}
\E\left[y_{t}|x_{t}\right]=f(x_{t}),\qquad x_{t}\in\mS \label{eq:regression}
\end{equation}
where $f:\mS\to\mathbb{R}$ is the
unknown regression function. Suppose that the conditional distribution of $ y_t $ given $ x_t=x $ is a Bernoulli distribution with mean $ f(x) $. We also assume that $ f $ is $1 $-Lipschitz continuous with respect to the Euclidean norm, i.e., 
\[
 |f(x)-f(x_0)|\leq  \vert x-x_0 \vert, \quad \forall x,x_0\in \mS.
 \]
Let $ \mathcal{F} $ be the collection of all $ 1 $-Lipschitz continuous function on $ \mathcal{X}$, i.e., 
\[
\mathcal{F}=\left\{ h|\text{\ensuremath{h} is a 1-Lipschitz function on \ensuremath{\mS}}\right\},
\] 
The goal is to estimate~$f$ given the observations $(x_{1},y_{1}),\ldots,(x_{T},y_{T})$ and the prior knowledge that $ f\in \mathcal{F} $. 

It is easy to verify that the above problem is a special case of the non-parametric regression problem considered in the work by \cite{stone1982optimal} (in particular, Example~2 therein).
Let $ \hat{f}_T $ denote an arbitrary (measurable) estimator of $ f $ based on the training samples $(x_{1},y_{1}),\ldots,(x_{T},y_{T})$.
By Theorem~1 in~\cite{stone1982optimal}, we have the following result: there exists a $ c>0 $ such that 
\begin{align}
\lim_{T\to\infty}\inf_{\hat{f}_{T}}\sup_{f\in\mF}\P\bigg(\big\Vert \hat{f}_{T}-f\big\Vert _{\infty}\ge c\Big(\frac{\log T}{T}\Big)^{\frac{1}{2+d}}\bigg)=1,
\end{align}
where infimum is over all possible estimators $ \hat{f}_T $.  Translating this result to the non-asymptotic regime, we obtain the
following theorem.
\begin{thm}
	\label{thm:regression_lower_bound}Under the above stated assumptions,
	for any number $\delta\in(0,1)$, there exits  $ c>0 $ and $T_{\delta}$
	such that 
	\[
	\inf_{\hat{f}_{T}}\sup_{f\in\mF}\P\bigg(\big\Vert \hat{f}_{T}-f\big\Vert _{\infty}\ge c\Big(\frac{\log T}{T}\Big)^{\frac{1}{2+d}}\bigg) \ge \delta, \qquad\text{for all \ensuremath{T\ge T_{\delta}}}.
	\]
\end{thm}

\medskip
\noindent{\bf Step 2. MDP with deterministic transitions.}
Consider a class of discrete-time discounted MDPs $ (\mS, \mA, \mathcal{P}, r, \gamma) $,
where
\begin{align*}
&\mS  =[0,1]^{d},\\
&\mA \text{ is finite},\\
&\text{for each} (x,a), \text{there exists a unique }x'\in \mS \text{such that} \mathcal{P}(x' |x,a)= 1,\\ 
&r(x,a)  =r(x)\text{ for all \ensuremath{a}},\\
&\gamma  =0.
\end{align*}
In words, the transition is deterministic, the expected reward
is independent of the action taken and the current state, and only immediate reward matters. 

Let $R_{t}$ be the observed reward at step $t$. We assume that given $x_t,$ the random variable $R_t$ is independent of $(x_{1},\ldots,x_{t-1})$, and follows a Bernoulli distribution  $\text{Bernoulli}\big(r(x_{t})\big).$
The expected reward function $r(\cdot)$ is assumed to be $1$-Lipschitz and bounded.
It is easy to see that for all $x\in\mS$, $a\in\mA$, 
\begin{align}
V^{*}(x) & =r(x). \label{eq:V_r_function}
\end{align}

\smallskip
\noindent{\bf Step 3. Reduction from regression to MDP.}
Given a non-parametric regression problem as described in Step $1$,
we may reduce it to the problem of estimating the value function $V^{*}$
of the MDP described in Step $2$. To do this, we set
\begin{align*}
r(x) & =f(x), \qquad\forall x\in\mS
\end{align*}
and
\begin{align*}
R_{t} & =y_{t},\qquad t=1,2,\ldots,T.
\end{align*}
In this case, it follows from equations~\eqref{eq:V_r_function} that the value function is given by $V^{*}=f$. Moreover, the expected reward function $r(\cdot)$ is $1$-Lipschitz, so the assumptions of the MDP in Step $ 2 $ are
satisfied. This reduction shows that the MDP problem is at least as hard as
the nonparametric regression problem, so a lower bound for the latter
is also a lower bound for the former. 

Applying Theorem~\ref{thm:regression_lower_bound} yields the following result:
for any number	$\delta\in(0,1)$, there exist some numbers $c>0$ and $T_{\delta}>0$, such that 
	\[
	\inf_{\hat{V}_T}\sup_{V^*\in \mF}\P\bigg[ \big\Vert \hat{V}_T-V^{*}\big\Vert_{\infty}\ge c\left(\frac{\log T}{T}\right)^{\frac{1}{2+d}} \bigg] \ge\delta, \qquad\text{for all \ensuremath{T\ge T_{\delta}}}.
	\]
	Consequently, for any reinforcement learning algorithm $\hat{V}_T$
	and any sufficiently small $\varepsilon>0$, there exists an MDP problem with deterministic transitions
	such that in order to achieve 
	\[
	\P\Big[\big\Vert \hat{V}_T-V^{*}\big\Vert_{\infty}<\varepsilon\Big]\ge 1-\delta,
	\]
	one must have 
	\[
	T\ge C'd\left(\frac{1}{\varepsilon}\right)^{2+d}\log\left(\frac{1}{\varepsilon}\right),
	\]
	where $C'>0$ is a constant. The statement of Proposition \ref{prop:lower_bound} follows by selecting $\delta = \frac12$.